\documentclass[entropy,article,accept,pdftex,moreauthors]{Definitions/mdpi}

\firstpage{1} 
\pubvolume{1}
\issuenum{1}
\articlenumber{0}
\pubyear{2024}
\copyrightyear{2024}
\externaleditor{Academic Editor: Ernestina Menasalvas}
\datereceived{29 June 2024 } 
\daterevised{7 August 2024 } 
\dateaccepted{20 August 2024  } 
\datepublished{ } 
\hreflink{https://doi.org/} 

\makeatletter
\let\c@lofdepth\relax
\let\c@lotdepth\relax
\makeatother
\usepackage{subfigure}
\usepackage[normalem]{ulem}
\usepackage{makecell}
\newcolumntype{"}{@{\hskip\tabcolsep\vrule width 1pt\hskip\tabcolsep}}

\usepackage{adjustbox}

\newlength{\figwidth}
\Title{Multifractal Hopscotch in \textit{Hopscotch} by Julio Cort\'azar}

\TitleCitation{Multifractal Hopscotch in \textit{Hopscotch} by Julio Cort\'azar}

\Author{Jakub Dec $^{1}$\orcidA{}, Michał Dolina $^{1}$\orcidB{}, Stanisław Drożdż $^{1,2,}$*\orcidC{}, Jarosław Kwapień $^{2,}$*\orcidD{} and Tomasz Stanisz $^{2}$\orcidE{}}

\AuthorNames{Jakub Dec, Michał Dolina, Stanisław Drożdż, Jarosław Kwapień and Tomasz Stanisz}

\AuthorCitation{Dec, J.; Dolina, M.; Drożdż, S.; Kwapień, J.; Stanisz, T.}

\address{%
$^{1}$ \quad Faculty of Computer Science and Telecommunications, Cracow University of Technology, \mbox{31-155 Krak\'ow, Poland;} jakub.dec@pk.edu.pl (J.D.); michal.dolina@pk.edu.pl (M.D.)\\
$^{2}$ \quad Complex Systems Theory Department, Institute of Nuclear Physics, Polish Academy of Sciences, \mbox{31-342 Krak\'ow, Poland;} tomasz.stanisz@ifj.edu.pl}

\corres{Correspondence: stanislaw.drozdz@ifj.edu.pl (S.D.); {jaroslaw.kwapien@ifj.edu.pl (J.K.)} 
}

\abstract{Punctuation is the main factor introducing correlations in natural language written texts and it crucially impacts their overall effectiveness, expressiveness, and readability. Punctuation marks at the end of sentences are of particular importance as their distribution can determine various complexity features of written natural language. Here, the sentence length variability (SLV) time series representing \textit{Hopscotch} by Julio Cort\'azar are subjected to quantitative analysis with an attempt to identify their distribution type, long-memory effects, and potential multiscale patterns. The analyzed novel is an important and innovative piece of literature whose essential property is freedom of movement between its building blocks given to a reader by the author. The statistical consequences of this freedom are closely investigated in both the original, Spanish version of the novel, and its translations into English and Polish. Clear evidence of rich multifractality in the SLV dynamics, with a left-sided asymmetry, however, is observed in all three language versions as well as in the versions with differently ordered chapters.}


\begin{document}

\section{Introduction}

One of the perspectives on research on natural language is the one derived from complexity science~\cite{HebertDufresne2024,KwapienJ-2012a}---as multiple traits identify natural language as a complex system~\cite{StaniszT-2024a}. Numerous methods widely used in studying complex systems, based on concepts originating in information theory~\cite{Debowski2020,Takahira2016,Montemurro2011}, time series analysis~\cite{AlvarezLacalle2006,Liu2023,Sanchez2023,Pawlowski1997,KosmidisK-2006a}, network science~\cite{Cancho2001,LiuH-2011a,Cong2014,WachsLopezGA-2016a,Kulig2017,Stanisz2019,Raducha2018} and the theory of power-law probability distributions~\cite{NarananS-1998a,Newman2005,AusloosM-2010a,Piantadosi2014}, have been used to grasp the various quantitative characteristics of natural language. Understanding the mechanisms behind such characteristics has the potential to improve the methods of natural language processing and generation. This is especially relevant today, given the heightened focus on large language models (LLMs)~\cite{ShanahanM-2023a,ZhaoWX-2023a}, incorporated in generative AI systems like OpenAI's ChatGPT, Microsoft's Bing Chat, and Google's Bard. One of the language properties recently investigated with statistical methods is the usage of punctuation in written language---it has been demonstrated that the distribution of word counts between consecutive punctuation marks in literary texts generally follows a discrete Weibull distribution~\cite{StaniszT-2023a,StaniszT-2024a,DecJ-2024a}. {What is interesting and significant, however, is the fact that for different languages the two parameters of this Weibull distribution are, to a large extent, selectively different and specific to a given language. Consistent with this fact, the punctuation distributions of texts after translation into another language are still governed by the Weibull distribution but with the parameters of the target language~\cite{StaniszT-2023a,StaniszT-2024a}.}
Interestingly, the patterns of sentence-ending punctuation marks, such as periods, exhibit much more flexibility and are less strictly bound by this distribution. These intervals, equivalent to sentence lengths, can therefore display a more diverse range of patterns.

Among the works considered important in the world literature, a book interesting from the aforementioned quantitative perspective is \textit{Hopscotch} by Julio Cortázar. At its core, this novel challenges traditional narrative structures and invites readers to engage with the text in a non-linear manner. The novel presents multiple reading options: one can either follow the conventional order of chapters or choose to explore the narrative via a suggested ``hopscotch'' order, jumping between chapters in a non-sequential fashion~\cite{BernsteinJS-1974a,SimpkinsS-1990a}. This experimental approach allows readers to experience the story in a unique, personalised way reflecting the author's desire to break away from the conventional storytelling. As a consequence of this free-form structure of the novel, the order of sentences and smaller textual constituents depends on a chosen sequence of chapters, which implies a strong impact on long-range correlations, both the linear and nonlinear ones. Since the novel comprises an eclectic mix of styles, including passages of stream-of-consciousness and experimental language play, the overall richness and depth of the narrative results in a convoluted, multilevel linguistic construct that, if transformed into numbers, is expected to reveal substantial complexity with self-similarity being one of its likely manifestations.

\section{Materials and Methods}

In its original Spanish version~\cite{HopscotchES}, \textit{Hopscotch} consists of 155 chapters that can be read in the printed order, the ``hopscotch'' order recommended by the author, in which the number of the subsequent chapter is explicitly given at the end of the previous one, or in any other order formed by permutating the chapters that is chosen by a reader. Because of such alternatives, all foreign-language translations preserve the structure of the novel and the author recommendations. To illustrate this point, along with the Spanish text, there are two translations considered as well: the English one~\cite{HopscotchEN} and the Polish one~\cite{HopscotchPL}. We transform the text in each language version into a sentence length variability (SLV) time series by counting words between each two consecutive sentence-ending punctuation marks (these can be periods, interrogation marks, exclamation marks, and other equivalent marks). {The related statistical data are collected in Table~\ref{tab::statistics}.} The SLV data reflect the evolution of the size of the main functional building blocks throughout the whole novel, which has been shown to be substantially informative~\cite{DrozdzS-2016a,StaniszT-2023a}.
\begin{table}[H]
\caption{Essential statistics on the SLV data considered in this study.}
\label{tab::statistics}
\newcolumntype{C}{>{\centering\arraybackslash}X}
\begin{tabularx}{\textwidth}{cCCC}
\toprule
 & \textbf{Spanish} & \textbf{English} & \textbf{Polish} \\ \midrule
Minimum sentence length (words) & 1 & 1 & 1 \\ \midrule
Maximum sentence length (words) & 850 & 860 & 681 \\ \midrule
Average sentence length (words) & 16.6 & 17.4 & 13.9 \\ \midrule
Minimum sentence length (characters) & 1 & 1 & 1 \\ \midrule
Maximum sentence length (characters) & 4037 & 4058 & 3730 \\ \midrule
Average sentence length (characters) & 77.3 & 77.7 & 74.9 \\ \midrule
Minimum chapter length (sentences) & 4 & 3 & 3 \\ \midrule
Maximum chapter length (sentences) & 1710 & 1720 & 1873 \\ \midrule
Average chapter length (sentences) & 108.0 & 109.3 & 125.5 \\ \bottomrule
\end{tabularx}
\end{table}

Punctuation may be viewed as a process of interrupting an otherwise ceaseless stream of words that improves understanding of the message and allows a reader to have necessary breaks. These functions make occurrences of the punctuation marks suitable for the survival analysis~\cite{MillerR-1997a}. It has already been documented in the literature that the inter-mark distances develop distributions that can be modelled by the discrete Weibull distribution~\cite{StaniszT-2023a,StaniszT-2024a}. This distribution is given by the following probability mass function~\cite{NakagawaT-1975a}:
\begin{equation}
f(k) = (1-p)^{k^{\beta}} - (1-p)^{(k+1)^{\beta}}, \quad p \in (0,1), \quad \beta>0
\label{eq::discrete.weibull.pmf}
\end{equation}
and the following cumulative distribution function:
\begin{equation}
\mathcal{F}(k)=1 - \left( 1-p \right)^{k^\beta}.
\label{eq::Weibull.CDF}
\end{equation}
The latter describes the probability that the random variable takes on a value greater than $k$. The discrete Weibull distribution can be considered as a generalisation of the geometric distribution with probability $p$ to which it conforms if $\beta = 1$, i.e., if the probability $\mathcal{F}(k)$ is constant in time. If $\beta > 1$, the probability increases with time while the opposite is true for $\beta < 1$. The discrete Weibull distribution is used in various fields, including survival analysis, weather forecasting, and study of textual data~\cite{JohnsonNL-1994a,MillerR-1997a,AltmannEG-2009a}. In the case of punctuation, $f(k)$ is the probability that a mark will occur exactly after $k$ words.

Self-similarity or a lack of characteristics scale is among the most significant properties of natural complex systems. From an empirical perspective, this property can manifest itself in a non-trivial temporal organisation of measurement outcomes formed as a time series. In particular, complexity is often associated with a cascade-like hierarchy of data points that reveal multiscaling, which makes practical methods of identifying this sort of organisation a valuable tool in complex systems research~\cite{JimenezJ-2000a}. Our experience so far allows us to consider the multifractal detrended fluctuation analysis (MFDFA) as the most reliable method in this respect~\cite{KantelhardtJ-2002a,OswiecimkaP-2006a}. This method constitutes a multiscale generalisation of an earlier, commonly used detrended fluctuation  analysis (DFA)~\cite{PengCK-1994a}. 

Here, the essential steps of the MFDFA algorithm are briefly sketched. Let $U=\{u_i\}_{i=1}^T$ be a time series of $T$ consecutive measurements of some observable $u$. We partition it into $M_s$ non-overlapping windows of length $s$ starting from both ends of $U$, which gives us $2 M_s$ such windows. In each window, we eliminate potential non-stationarity of the signal by applying a detrending procedure to an integrated signal (the so-called profile) $X=\{x_i\}_{i=1}^s$ whose elements have the following form:
\begin{equation}
x_i = \sum_{j=1}^i u_j.
\end{equation}
The detrending is carried out with the help of a polynomial $P^{(m)}$ of order $m$ (we use $m=2$ throughout this study) that is best-fitted to $X$ in each window $\nu=0,\ldots,2 M_s-1$ and the variance of the resulting detrended signal is then calculated:
\begin{equation}
f^2(\nu,s) = {1 \over s} \sum_{i=1}^s (x_i - P^{(m)}(i))^2.
\label{eq::variance}
\end{equation}
A family of fluctuation functions of order $q$ is defined in the next step based on the average variance across all the windows:
\begin{equation}
F_q(s) = \left\{ {1 \over 2M_s} \sum_{\nu=0}^{2M_s-1} \left[ f^2(\nu,s) \right]^{q/2} \right\}^{1/q},
\label{eq::fluctuation.functions}
\end{equation}
where $q$ is a real number. The functions $F_q(s)$ have to be calculated for a number of different values of scale $s$ and index $q$. Typically, minimum $s$ is chosen above the length of the longest sequence of constant values of $U$ and maximum $s$ is $T/5$. In contrast, there is no typical range of $q$. This index can be viewed as being related to the moments of the signal, so the extreme values cannot be chosen as too large for a time series with heavy tails.

If the fluctuation functions depend on $s$ as power laws:
\begin{equation}
F_q(s) \sim s^{h(q)}
\label{eq::scaling}
\end{equation}
for a number of different choices of $q$, it indicates that the time series under study is either monofractal (when $h(q)$ is constant in $q$) or multifractal otherwise. The function $h(q)$ is called the generalised Hurst exponent, because for $q=2$, $h(q)=H$, where $H$ is the standard Hurst exponent~\cite{HurstHE-1951a,HeneghanC-2000a}. In terms of visualisation, fractal $F_q(s)$ form linear plots on double logarithmic plots. A convenient way of presenting the multifractal property of data is singularity spectrum $f(\alpha)$. It can be derived from $h(q)$ by the following formulae:
\begin{eqnarray}
\nonumber
\alpha = h(q) + qh'(q),\\
f(\alpha) = q \left[\alpha-h(q)\right] + 1,
\label{eq::singularity.spectrum}
\end{eqnarray}
where $\alpha$ is a measure of data-point singularity equivalent to the H\"older exponent and $f(\alpha)$ is the Legendre transform of $h(q)$. The function $f(\alpha)$ can be interpreted geometrically as a fractal dimension of the subset of the whole data set with the H\"older exponent equal to $\alpha$~\cite{HalseyTC-1986a}. For a monofractal time series the pair $(\alpha,f(\alpha))$ is a single point, while for a multifractal one it is a parabola with shoulders pointing down. The usefulness of this representation comes from a fact that the broader the singularity spectrum is, the richer is the multifractality of the time series, which can be viewed as a measure of time series complexity. Sometimes the parabola of $f(\alpha)$ is distorted and asymmetric, which indicates that the data points of different amplitude have different scaling properties~\mbox{\cite{OhashiK-2000a,CaoG-2013a,DrozdzS-2015a,GomezGomezJ-2021a}.} Alternatively, the scaling type may be expressed in terms of the multifractal spectrum $\tau(q)$ defined by
\begin{equation}
\tau(q) = qh(q)-1.
\end{equation}
For monofractal time series, $\tau(q)$ depends linearly on $q$ (because $h(q)$ is constant then), while it is nonlinear for multifractal ones, which allows one for an easy detection of multiscaling.

\section{Results and Discussion}

\subsection{Linear Correlations}

\begin{figure}[H]

\includegraphics[clip, width=0.49\linewidth]{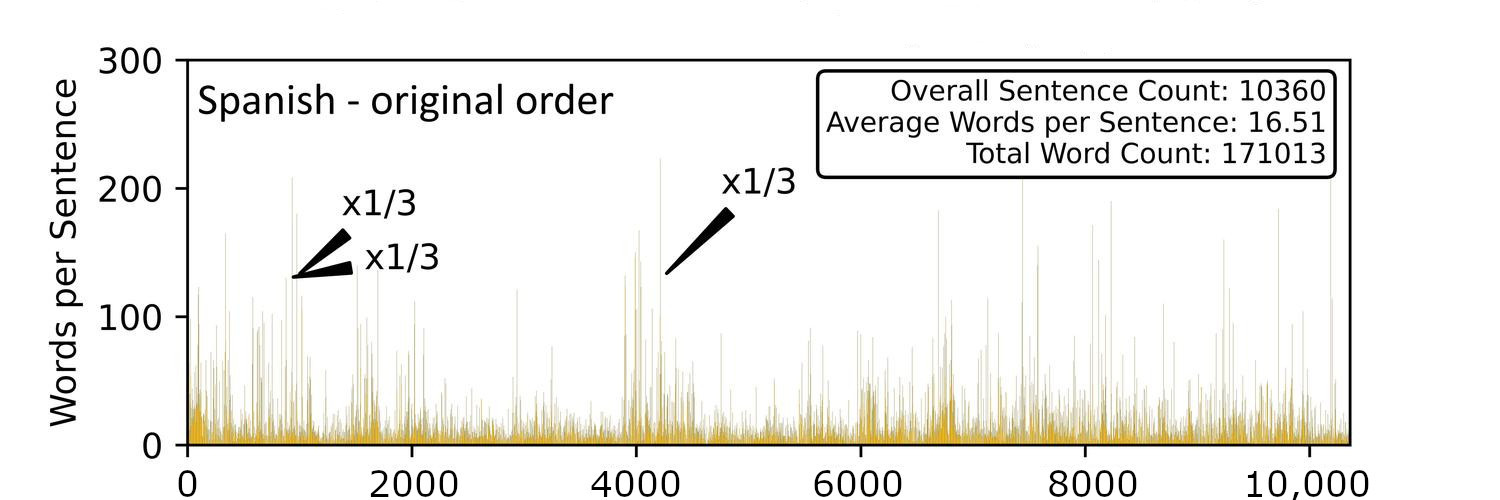}
\includegraphics[clip, width=0.49\linewidth]{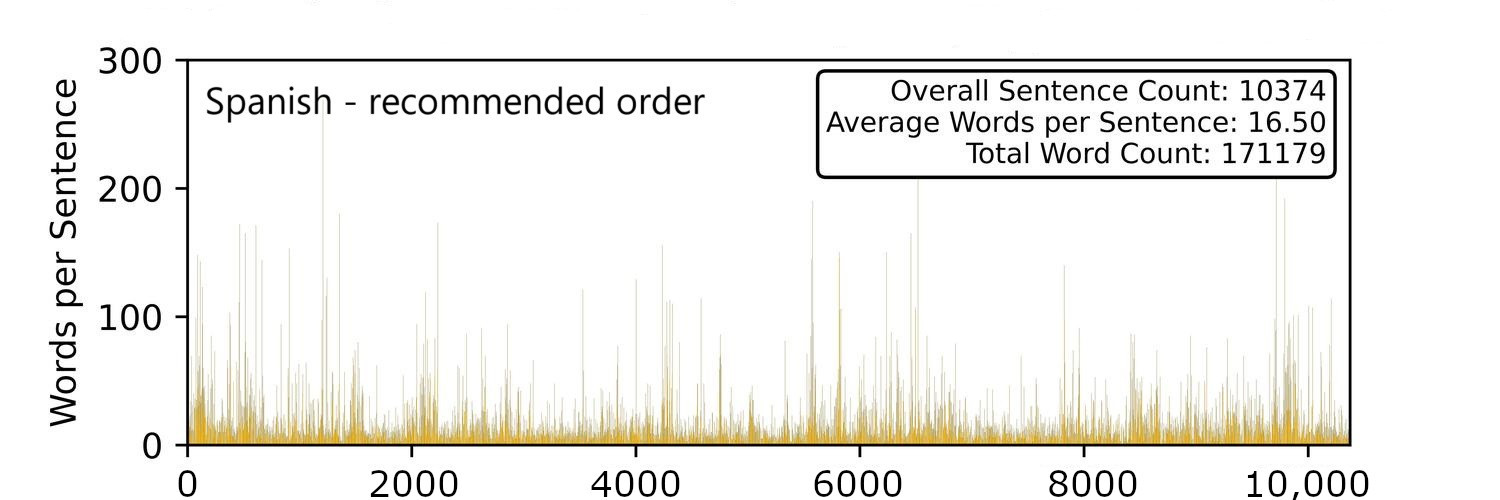}
\includegraphics[clip, width=0.49\linewidth]{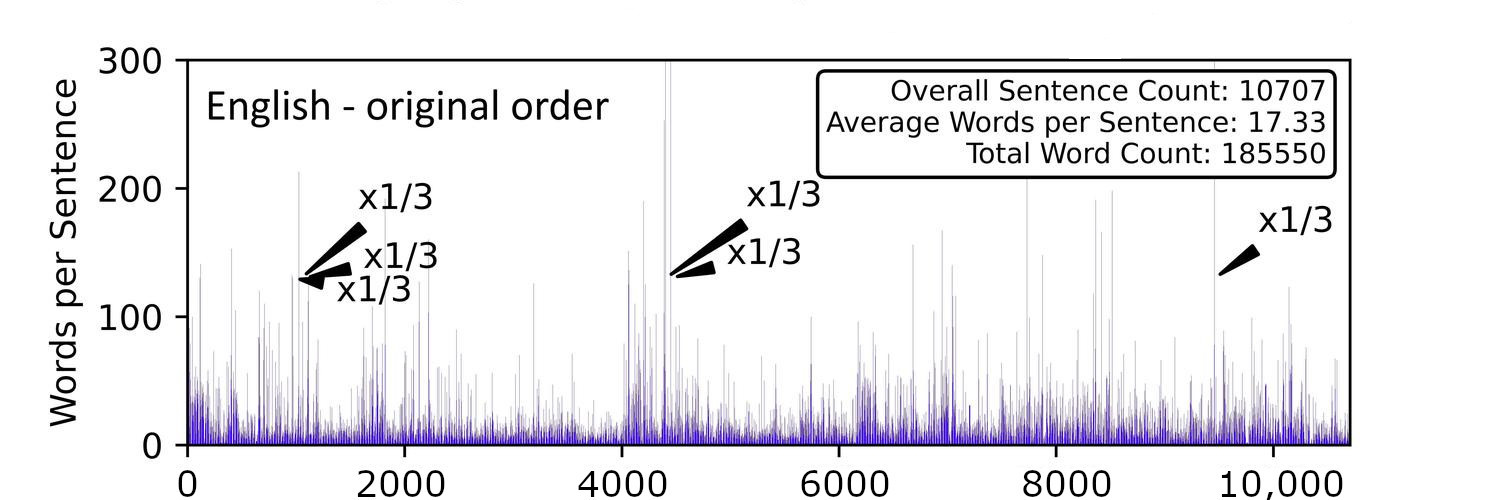}
\includegraphics[clip, width=0.49\linewidth]{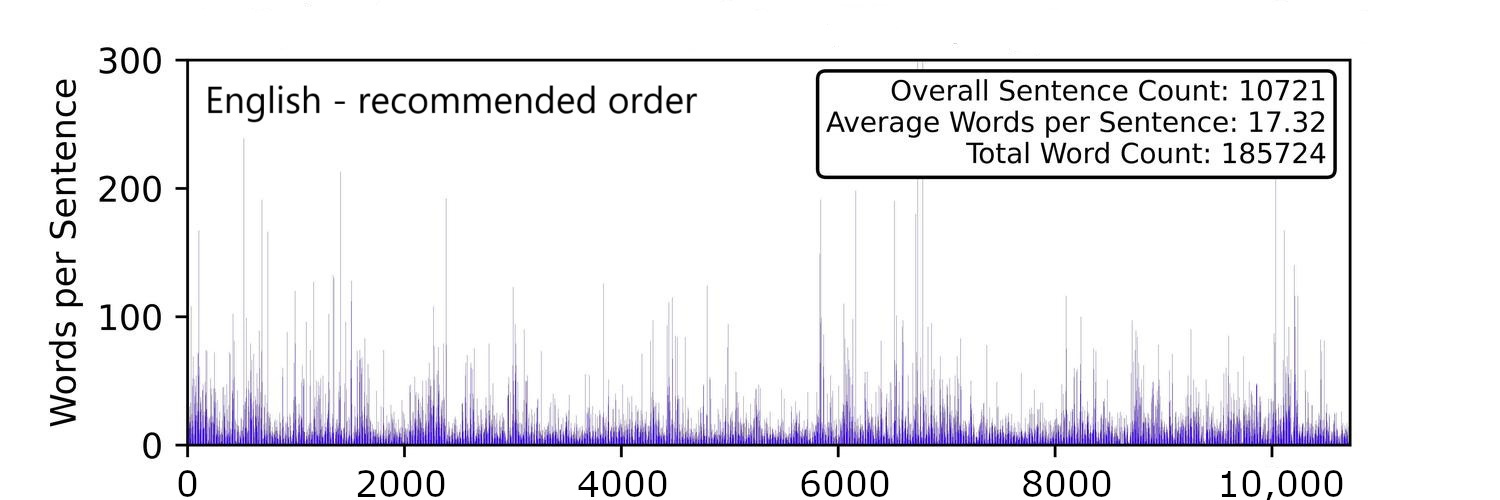}
\includegraphics[clip, width=0.49\linewidth]{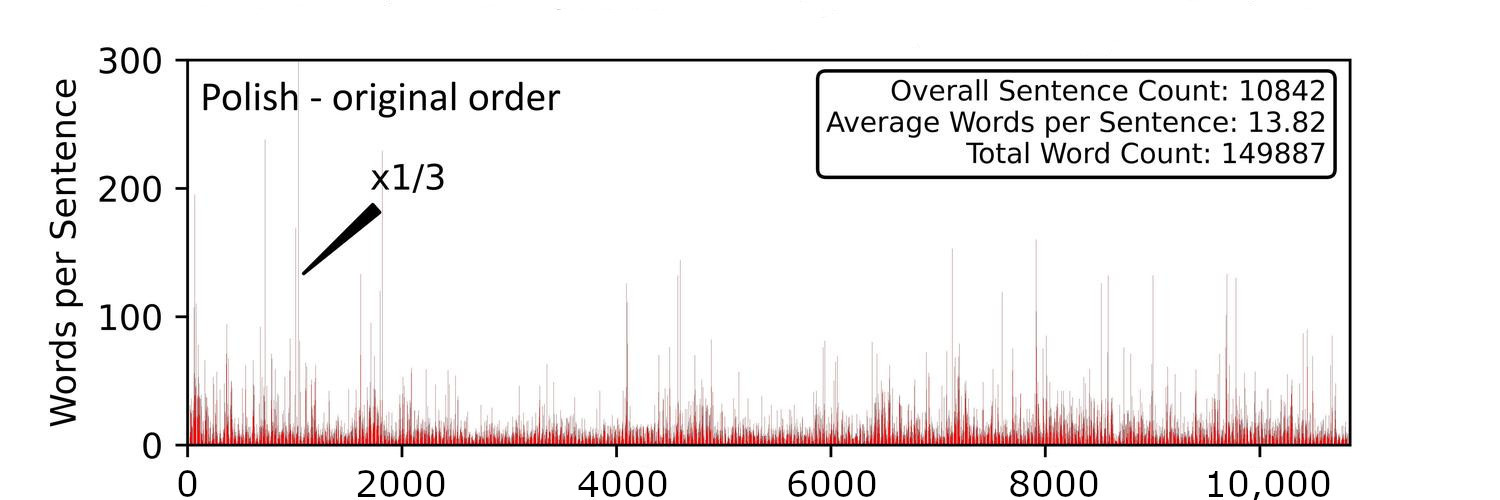}\hspace{2mm}
\includegraphics[clip, width=0.49\linewidth]{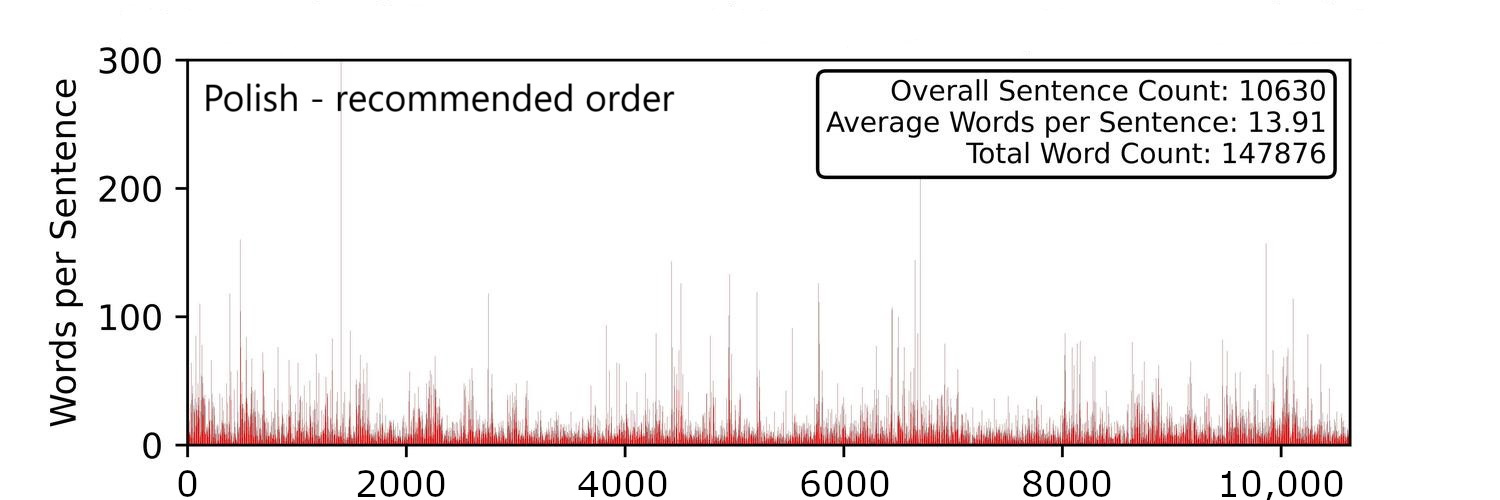}
\caption{Time series of sentence length variability (SLV) measured in words for the original Spanish text of \textit{Hopscotch} (\textbf{top}) and its translations into English (\textbf{middle}) and Polish (\textbf{bottom}). In order to ensure sufficient readability and preserve the vertical scale in all panels, the bars representing a few sentences with excessive length in each text have been rescaled by a factor of 1/3 (black arrows). Time series of SLV are presented in alternative orders of chapters: the printed order (\textbf{left}) and the order recommended by the author (\textbf{right}).}
\label{fig::slv.time.series}
\end{figure}

It is instructive to look into data structure before any further step is made. In the left column of Figure~\ref{fig::slv.time.series}, time series of SLV in three different language versions of the text are shown, each of them representing an unsigned process with a heavy tail. The SLV time series representing the text with two alternative chapter orders, the one recommended by the author and a completely random one, are shown in the right column (top and middle panels) of the same figure. In both cases,  there is a clear data clustering observed (long sentences tend to group with the long ones, while small sentences tend to group with the small ones), the time series are characterised by memory. This by no means differs from other pieces of literature where temporal memory is naturally present. It stems from the fact that sentence lengths are often connected with writing style and certain features of narrative, which usually last for some paragraphs or pages (for instance, longer sentences are often used in descriptions and slow-paced parts of texts, while short sentences are more typical for fast-paced parts and dialogues). This property vanishes if the sentences are shuffled instead of the chapters.

\begin{figure}[H]

\includegraphics[scale=0.4]{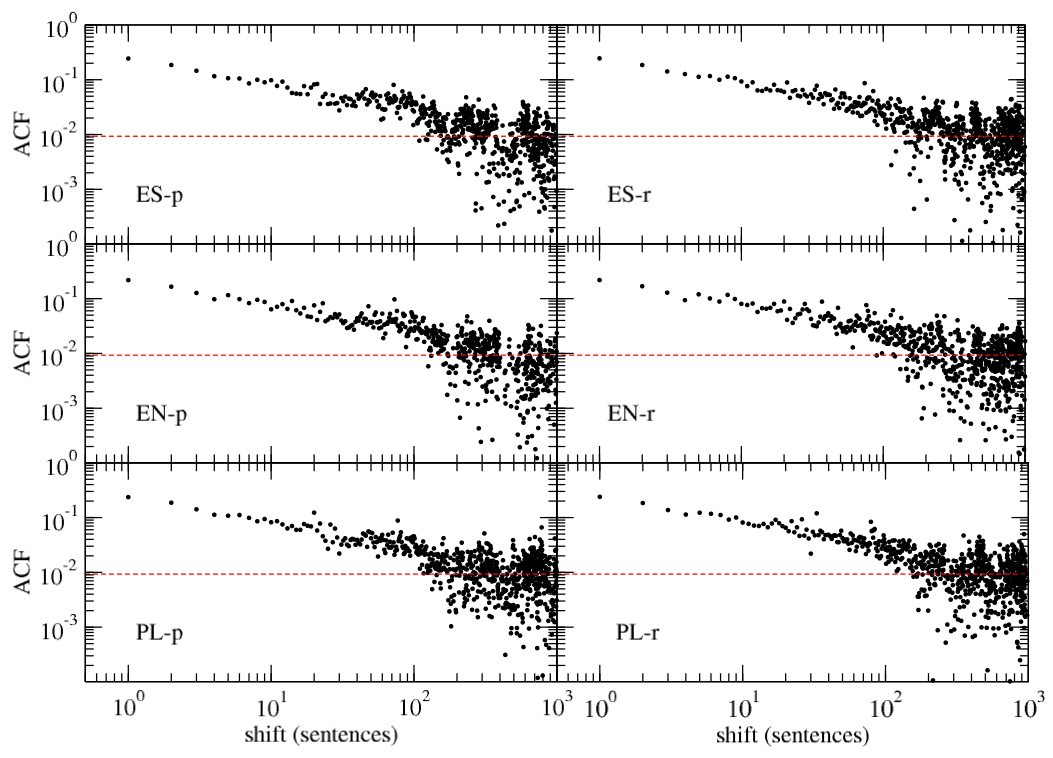}
\caption{Autocorrelation function for SLV time series for two chapter orders: the printed order (``-p'', \textbf{left}) and the recommended order (``-r'', \textbf{right}), and for three languages: Spanish (ES, \textbf{top}), English (EN, \textbf{middle}), and Polish (PL, \textbf{bottom}). Noise level is indicated by a red dashed line in each panel. Note the double logarithmic scale of the plots.}
\label{fig::slv.acf}
\end{figure}

Plots of the Pearson autocorrelation function (ACF) for the two authorial chapter orders are presented in Figure~\ref{fig::slv.acf}. ACF shows statistically meaningful values over a range of up to 200--400 consecutive sentences, after which it reaches noise level in both chapter orders. A trace of power-law decay of ACF can be noticed because of the double-log scale of the plots. Roughly the same picture is obtained if a random permutation of chapters is considered. This means that there is no difference between the proposed chapter orders and any other order that can be preferred by a reader (it must be noted, however, that it is impossible to consider all possible orderings as we deal with $155! \approx 10^{273}$ possible permutations here). As it might be expected, the sentence-level randomisation kills any genuine structure of SLV.

\unskip

\subsection{The Weibull Analysis}

\begin{figure}[H]

\includegraphics[width=0.49\linewidth]{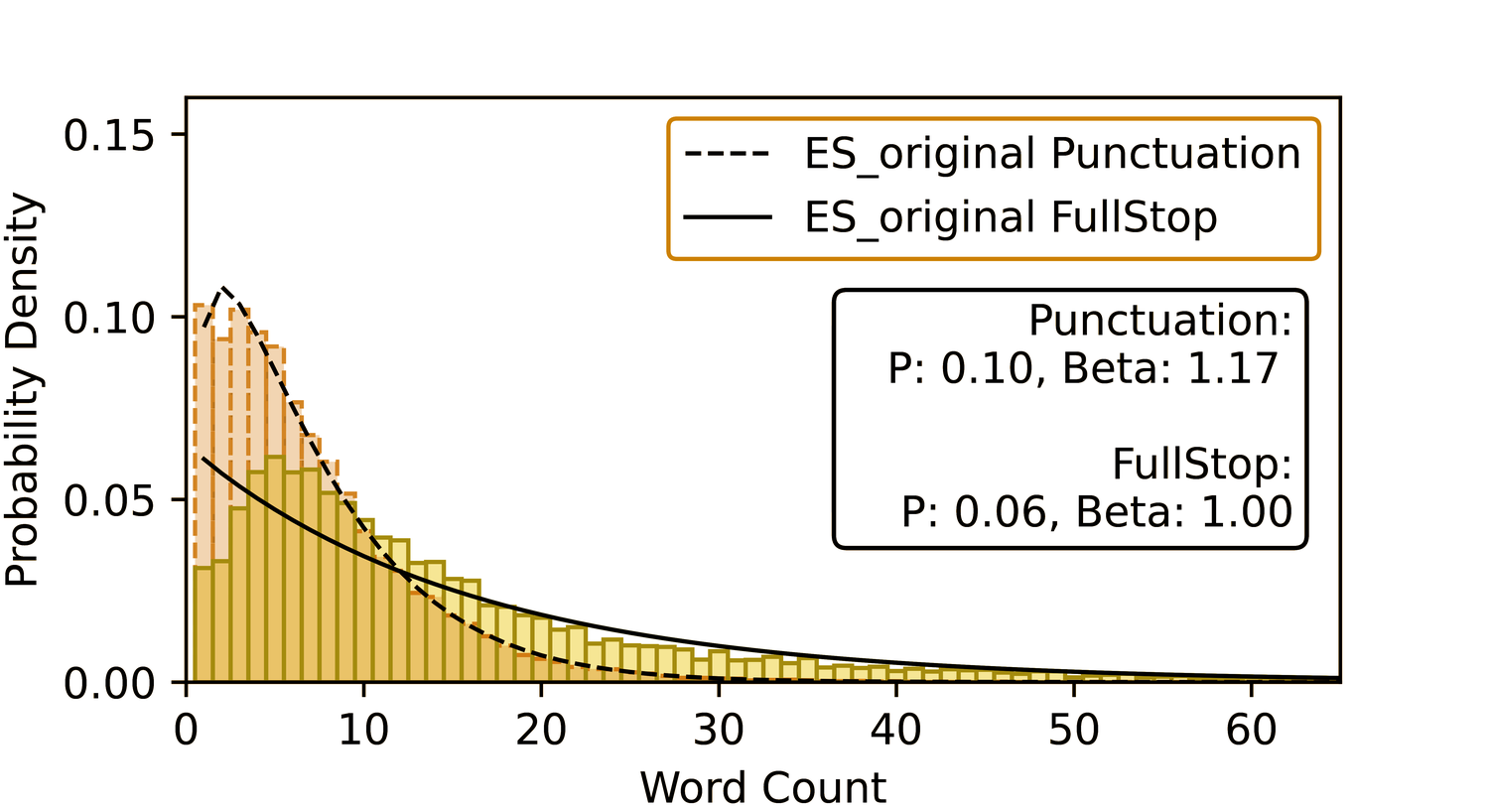}
\includegraphics[width=0.49\linewidth]{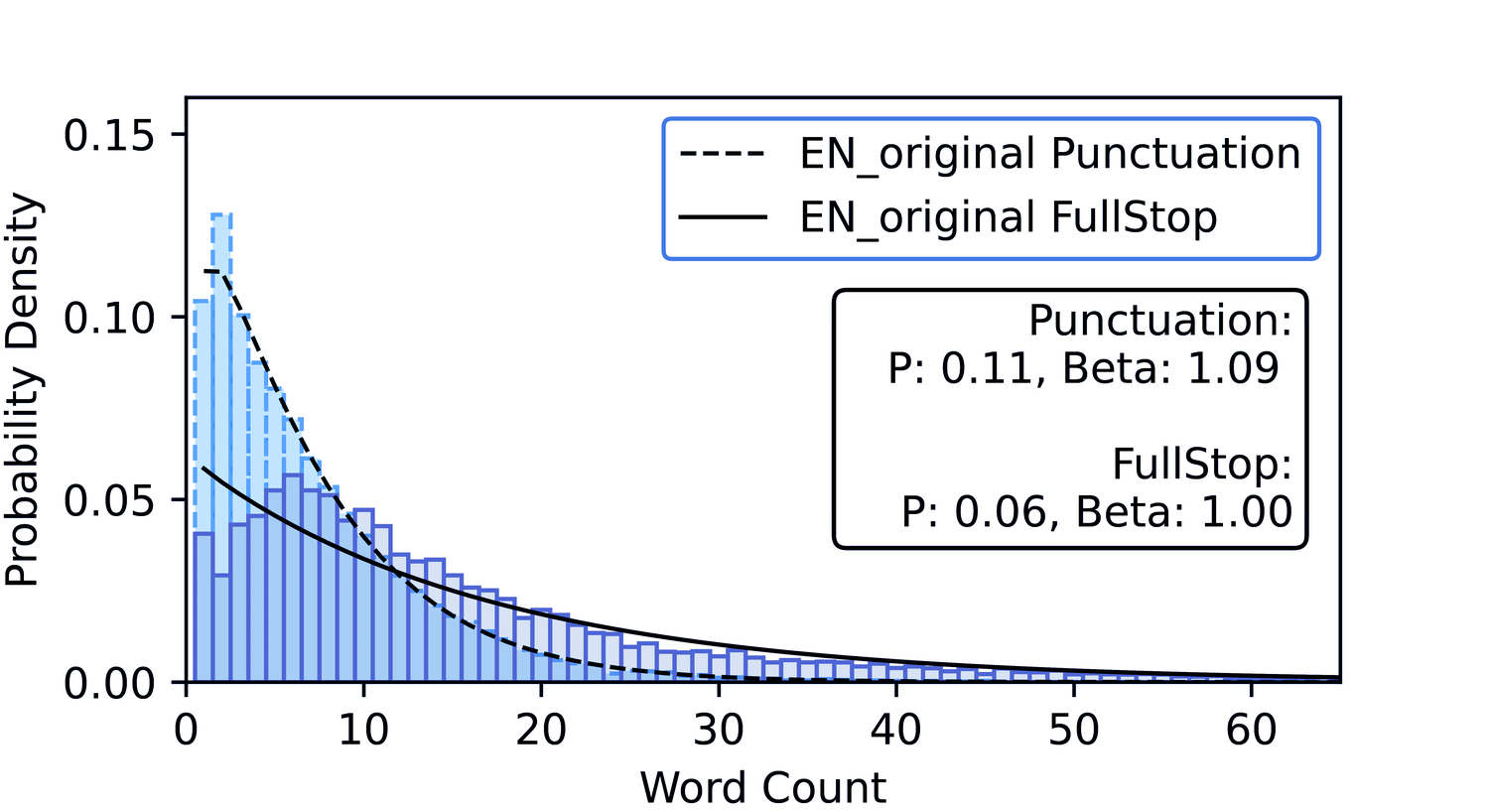}
\includegraphics[width=0.49\linewidth]{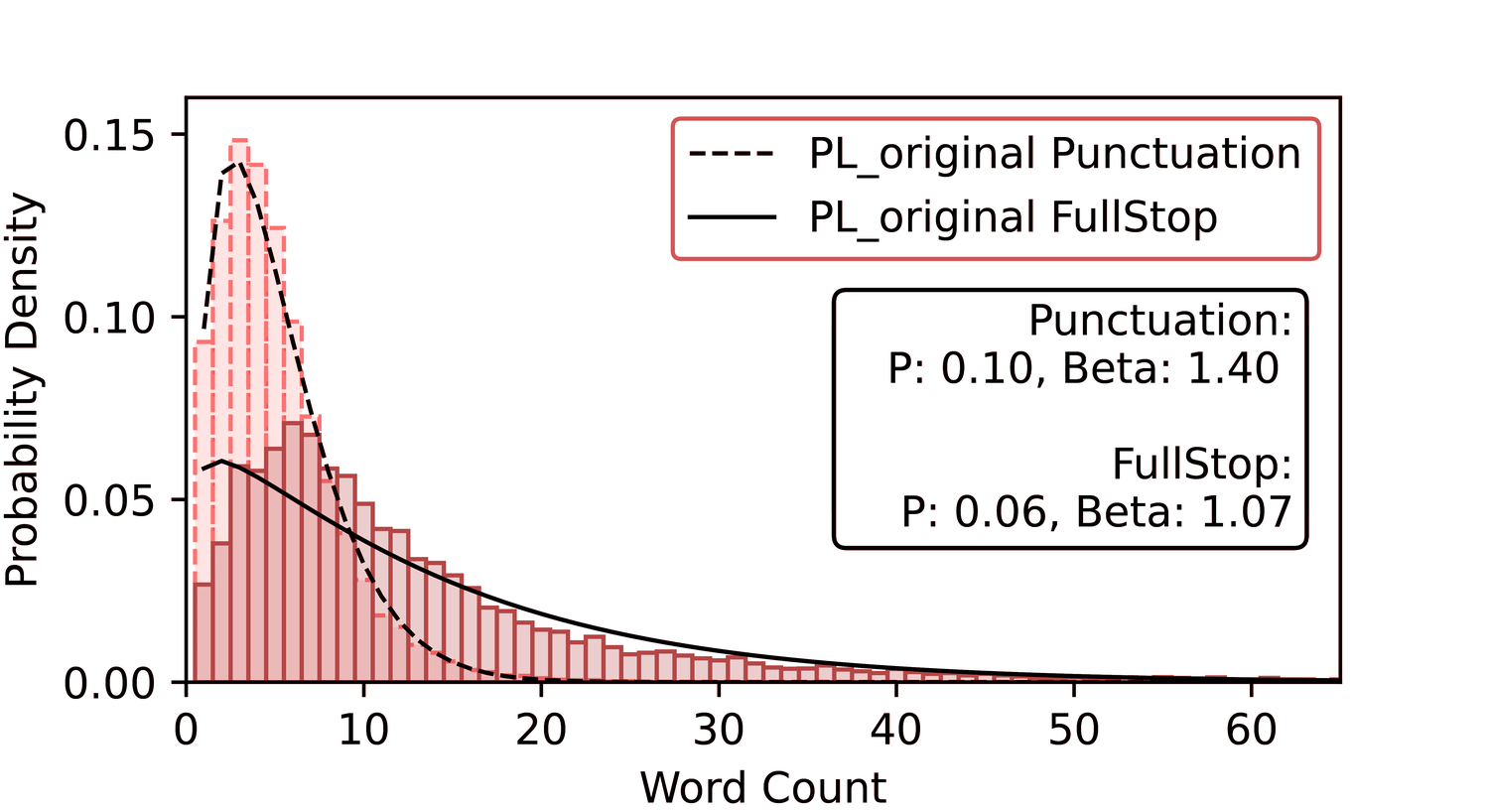}
\caption{Histograms of the SLV time series (dark colour) for three different language versions of \textit{Hopscotch}: Spanish (\textbf{top left}), English (\textbf{top right}), and Polish (\textbf{bottom}) together with the histograms of the distances (in words) between any two consecutive punctuation marks (both the sentence-ending ones and the intra-sentence ones, light colour). Discrete Weibull distributions that have been least-square fitted to the histograms of both types are also shown in each panel (dashed and solid lines, respectively). In each case, values of the $p$ and $\beta$ parameters of the fits are explicitly given in legend boxes. (Insets) Histogram of the SLV time series with the fitted discrete Weibull distribution presented on the half-logarithmic scale.}
\label{fig::slv.weibull}
\end{figure}

For all three language versions of \textit{Hopscotch}, histograms for the respective SLV time series have been calculated; they are shown in Figure~\ref{fig::slv.weibull} (dark colours) together with the best-fitted discrete Weibull distribution defined by Equation~(\ref{eq::discrete.weibull.pmf}). It is evident that there is no agreement between theory and the empirical distribution in each case---the result that matches results of similar analyses of different texts already reported in the literature~\cite{StaniszT-2023a,StaniszT-2024a}. A much better agreement between the two can be seen for a different type of time series (punctuation mark distance variability, PMDV), in which distances between consecutive punctuation marks have been considered without restricting the analysis to sentence endings. Thus, if also such marks as comma, colon, semicolon, dash, etc. are included in the study, the discrete Weibull distribution becomes a better model for the data (see light-colour histograms in Figure~\ref{fig::slv.weibull}). For all the language versions, the values of $\beta$ for the PMDV time series exceed 1, which can be viewed as the normal case for written language, in which the probability of using a punctuation mark after a given number of words since the previous instance of writing a mark (the hazard function) increases with this number. It distinguishes \textit{Hopscotch} from \textit{Finnegans Wake} by James Joyce, where an abnormal value $\beta<1$ has been found~\cite{StaniszT-2024b}. In contrast, for the SLV data, $\beta \approx 1$ has been obtained for two out of three languages; it corresponds to the exponential distribution of such data. This is better seen if the plots are semi-logarithmic (see the insets in Figure~\ref{fig::slv.weibull}), where the exponential decay of the histogram tails and the corresponding fits are represented by straight lines.
\vspace{-9pt}

\subsection{Generalised Hurst Exponents}

\begin{figure}[H]

\includegraphics[trim={0cm 0 0 3.75cm}, clip, width=0.49\linewidth]{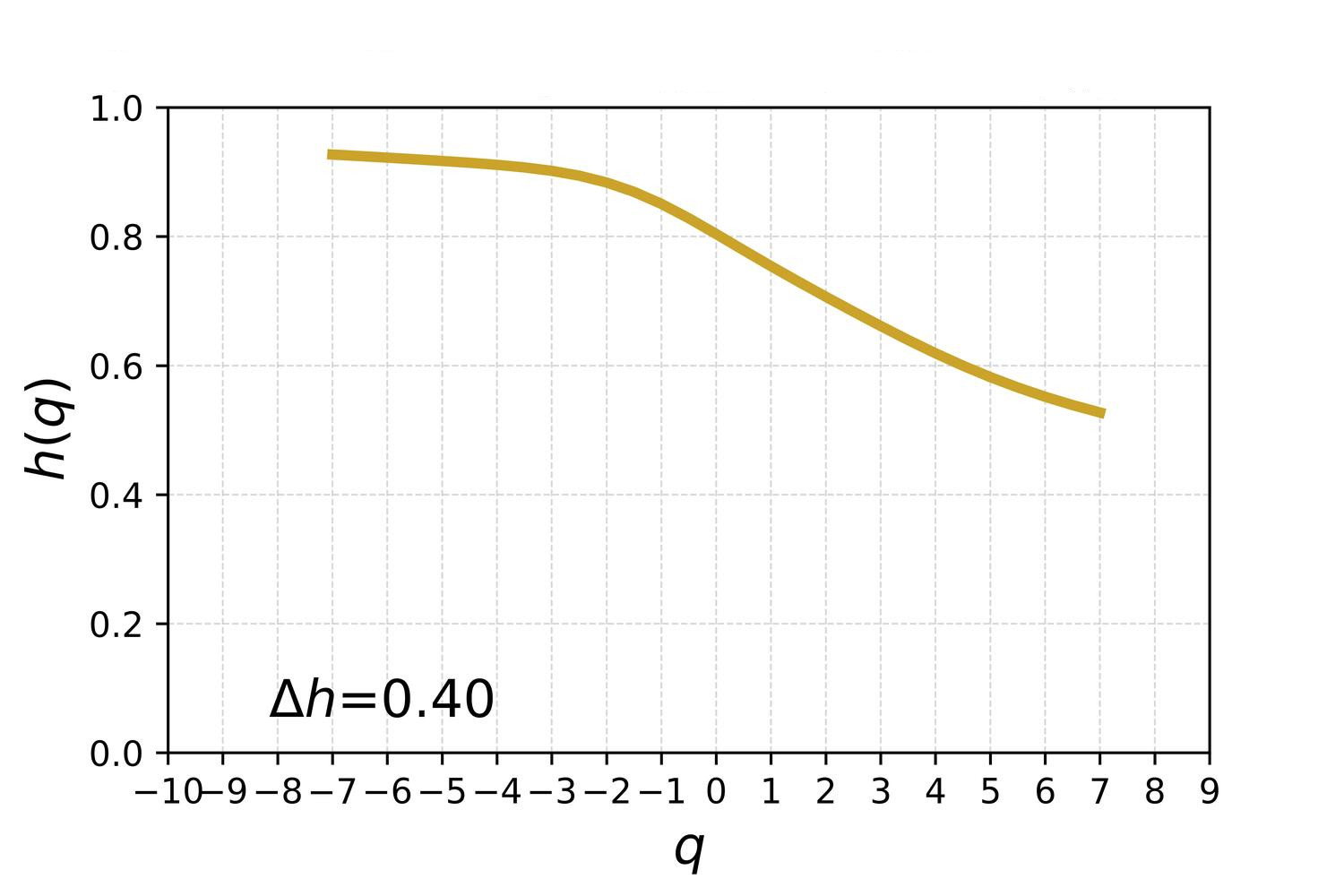}
\includegraphics[trim={0cm 0 0 3.75cm}, clip, width=0.49\linewidth]{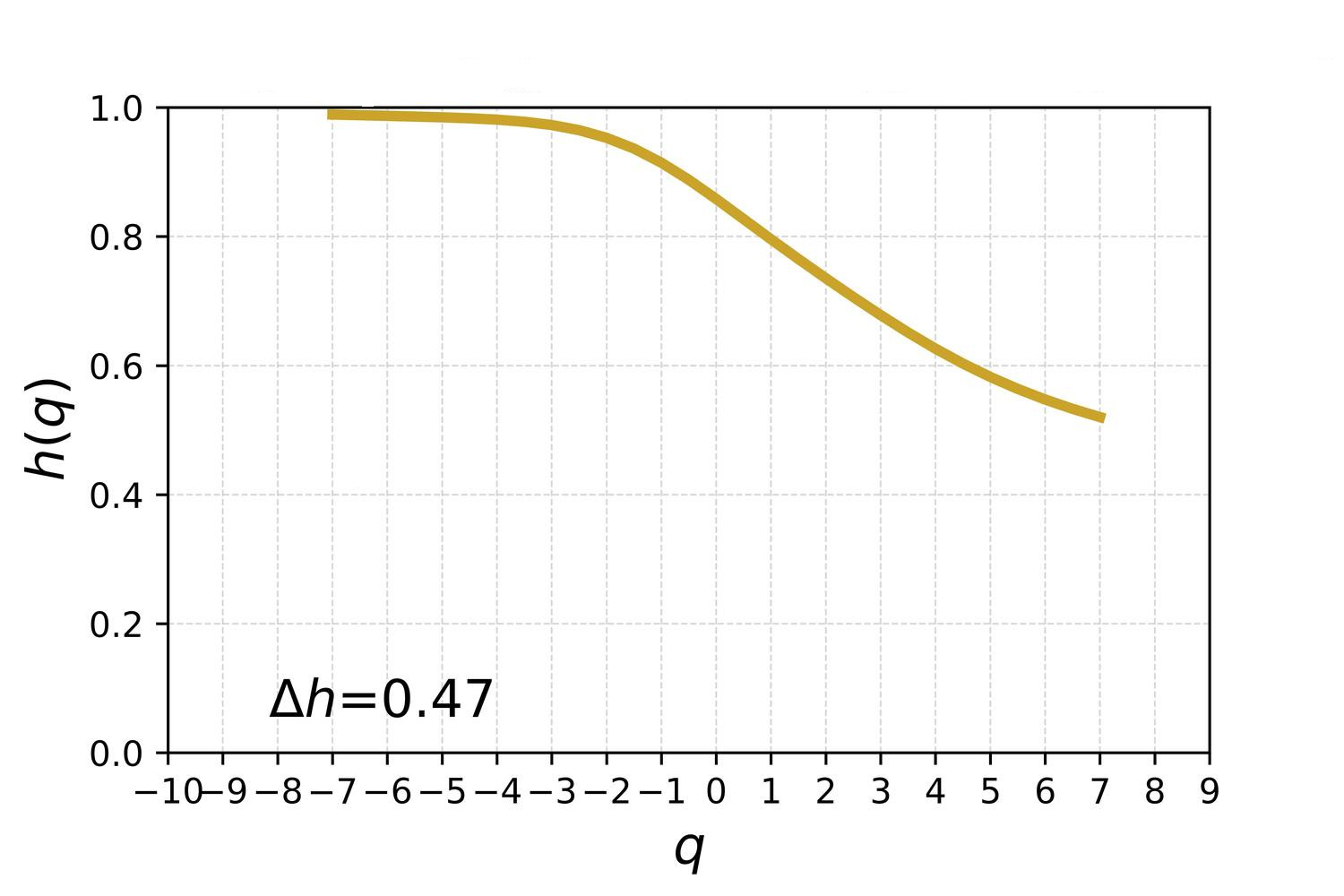}
\includegraphics[trim={0cm 0 0 3.75cm}, clip, width=0.49\linewidth]{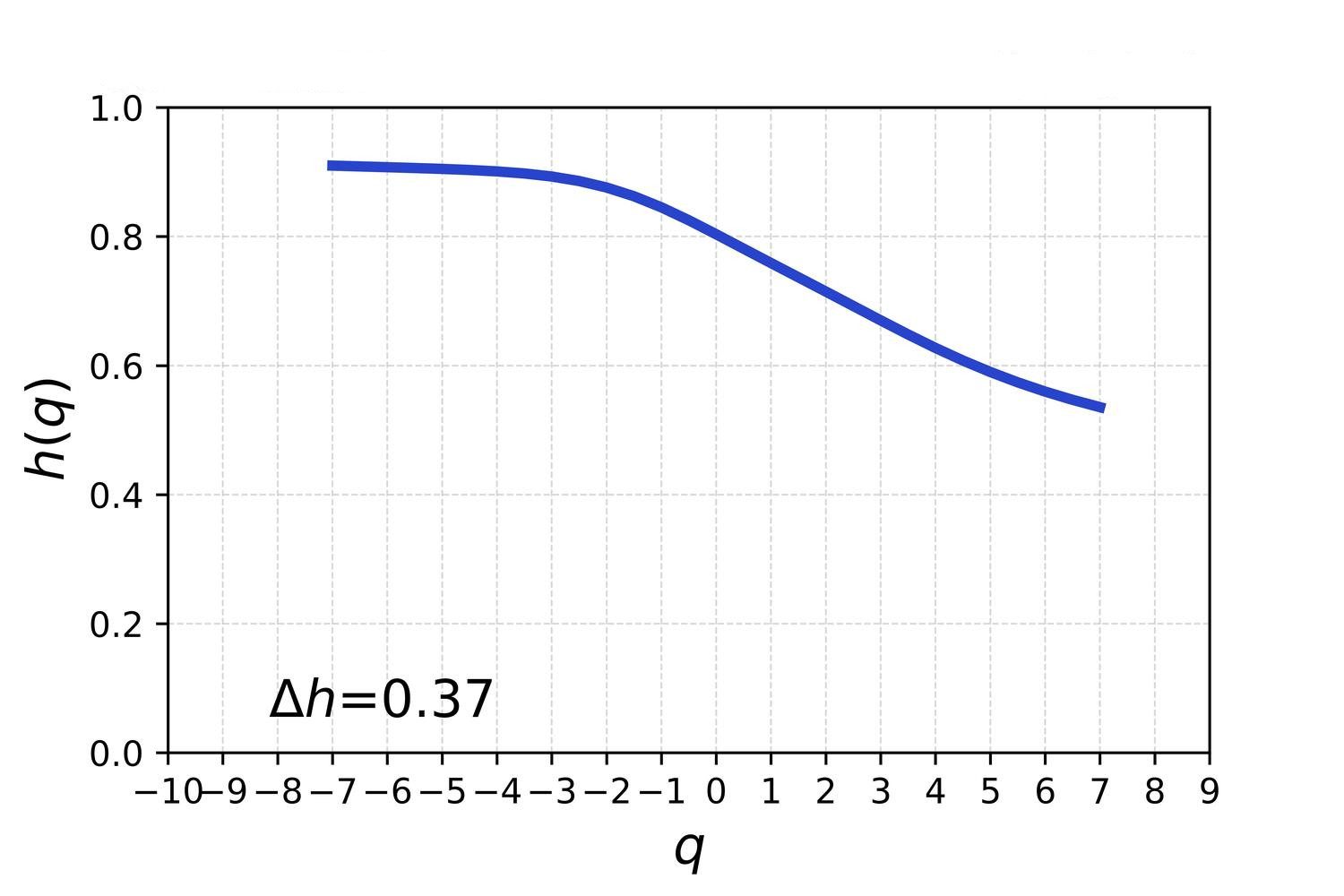}
\includegraphics[trim={0cm 0 0 3.75cm}, clip, width=0.49\linewidth]{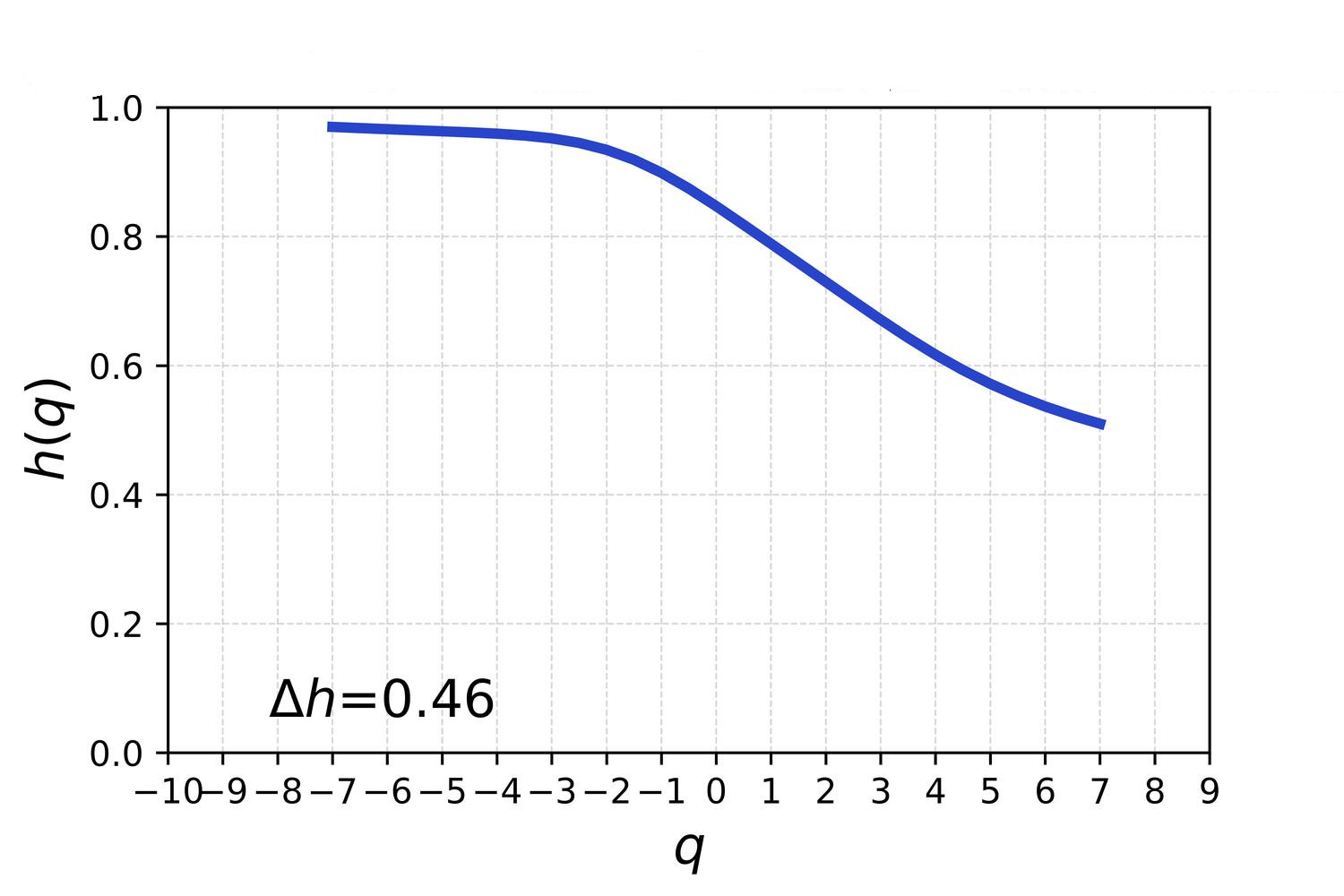}
\includegraphics[trim={0cm 0 0 3.75cm}, clip, width=0.49\linewidth]{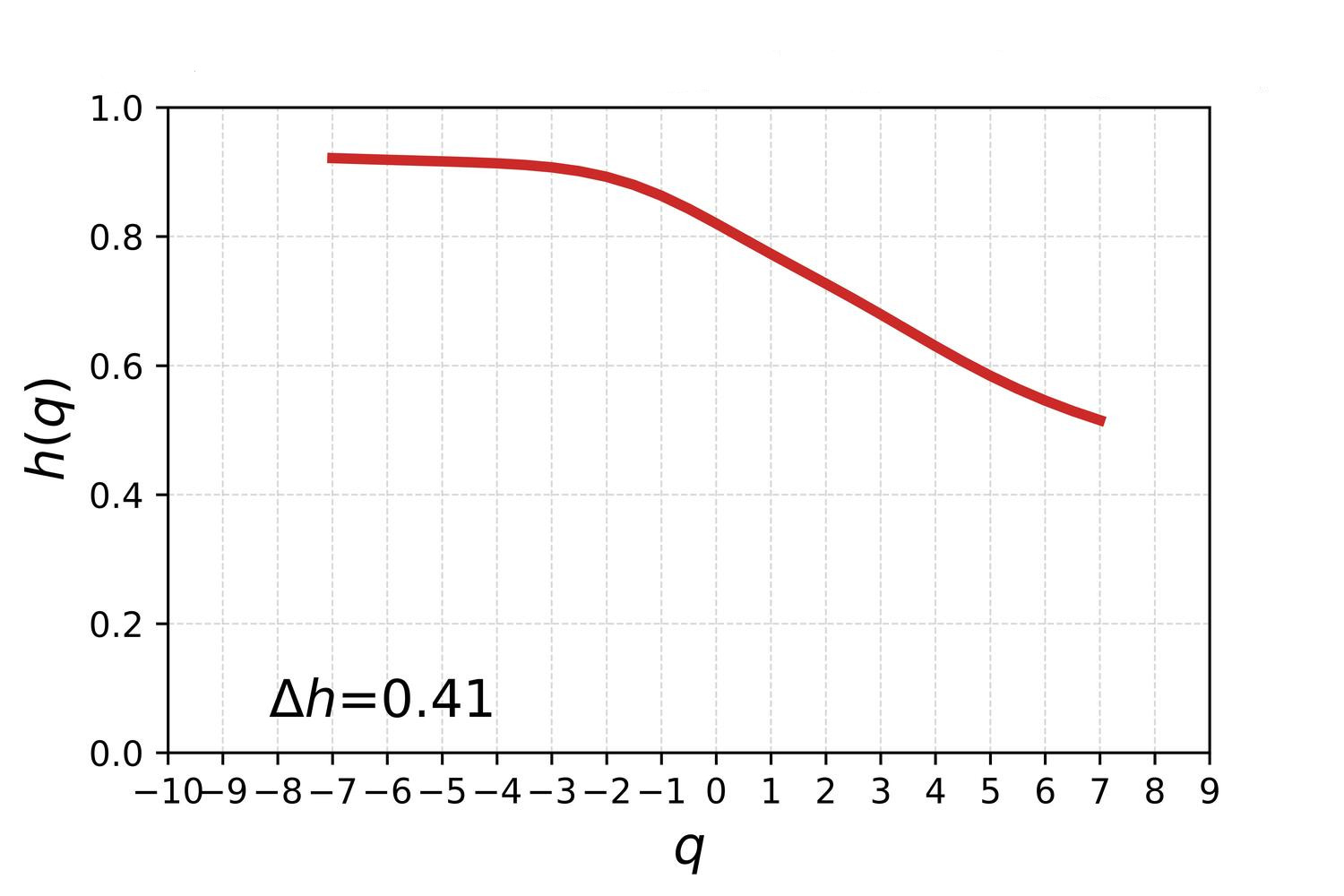}
\includegraphics[trim={0cm 0 0 3.75cm}, clip, width=0.49\linewidth]{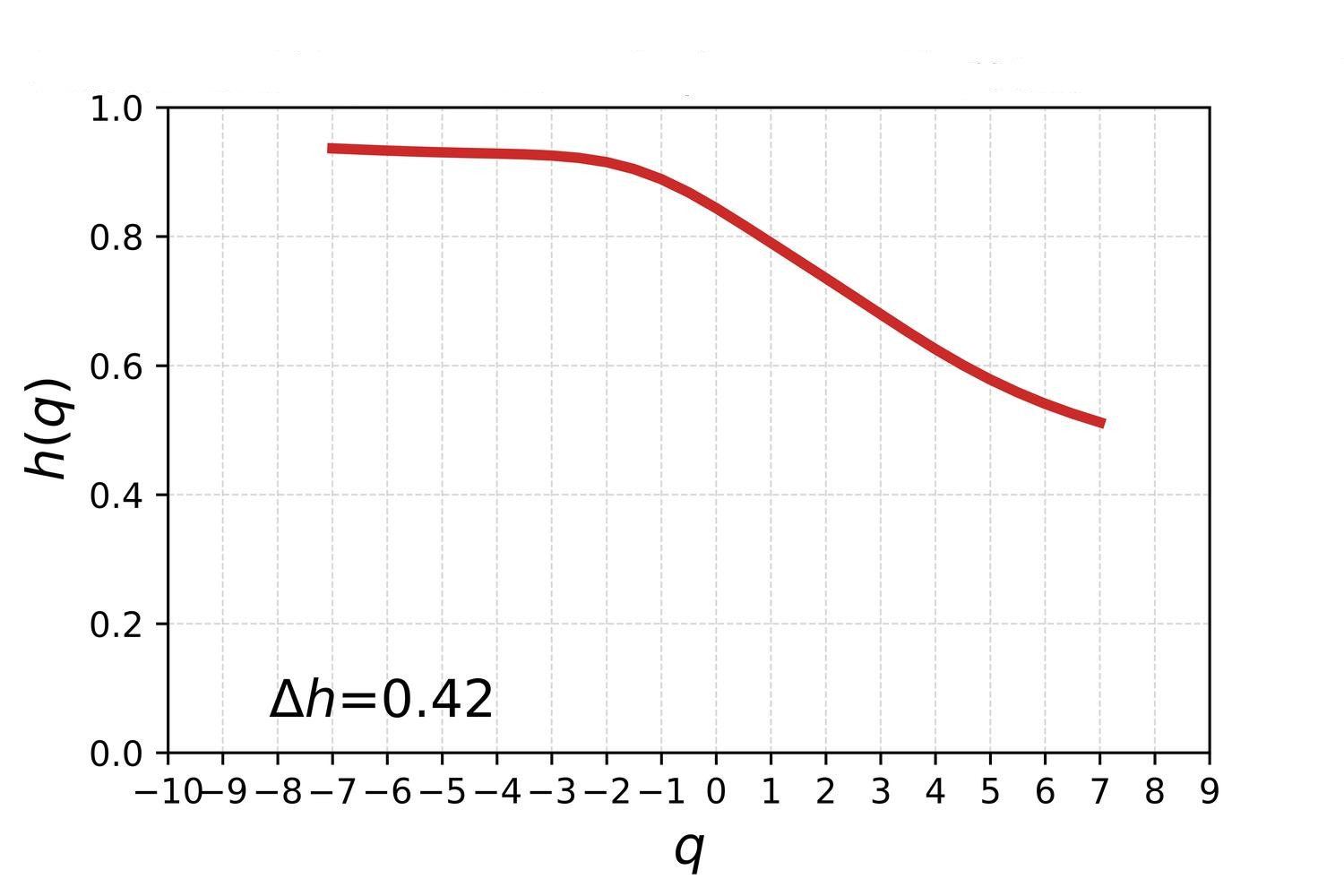}
\caption{The generalised Hurst exponent $h(q)$ for the SLV time series representing \textit{Hopscotch} in different languages: Spanish (\textbf{top}), English (\textbf{middle}), and Polish (\textbf{bottom}). The printed order (\textbf{left}) and the recommended order (\textbf{right}) of chapters are shown separately. The difference between extreme values of $h(q)$ for $-7 \le q \le 7$ is given by $\Delta h$ in each case.}
\label{fig::slv.hurst}
\end{figure}

The classic Hurst analysis of time series allows one to quantify linear autocorrelation, as it expresses how the observed value span depends on the length of observation~\cite{HurstHE-1951a}. In its generalised version $h(q)$, it allows to detect differences in memory effects between fluctuations of different magnitudes~\cite{MonjoR-2024a}. It naturally enters the multiscale analysis through Equation~(\ref{eq::scaling}). $h(q)$ is a non-increasing function of its argument with the special case being a constant function $h(q)=H$ for all $q$s. In a practical situation, the range of considered values of $q$ must be limited and dependent on the probability distribution function of the underlying stochastic process. Typically, the reference is that the $q$th moment of this distribution has to exist. In the present case, the related probability mass function is close to exponential, therefore no specific theoretical limit for $q$ has to be applied. Because of this freedom, the parameter $q$ has been limited to $-7 \le q \le 7$, which is a reasonable compromise between the range width, which should be as large as possible, and the available sample size. Figure~\ref{fig::slv.hurst} shows the generalised Hurst exponents for the three languages and the two authorial chapter orders. In each case $h(q)$ is a decreasing function which indicates that the time series are heterogeneous in their scaling behaviour for different $q$s. The larger the variability of $\Delta h = h(q_{\rm max}) - h(q_{\rm min})$ is, the more heterogeneous they are. However, as $0.37 \le \Delta h \le 0.47$ the difference between the cases is more quantitative than qualitative. It is not surprising, because, on the one hand, by changing the order of chapters, one effectively preserves the autocorrelation range up to the average number of sentences per chapter and, on the other hand, good translations into foreign languages are expected to preserve the structure of the original language version as closely as possible.
\vspace{-6pt}

\subsection{Multifractal Analysis}

\begin{figure}[H]

\begin{adjustwidth}{-\extralength}{0cm}
\centering
\renewcommand{\tabularxcolumn}[1]{m{#1}}
\setlength\tabcolsep{1pt}
\begin{tabularx}{\textwidth}{| >{\centering\arraybackslash}m{0.75\textwidth} | >{\centering\arraybackslash}m{0.25\textwidth} |}
\hline
\multirow{3}{*}{
\adjustbox{valign=t}{\includegraphics[trim={0cm 0 0 3cm}, clip, width=1.07\linewidth]{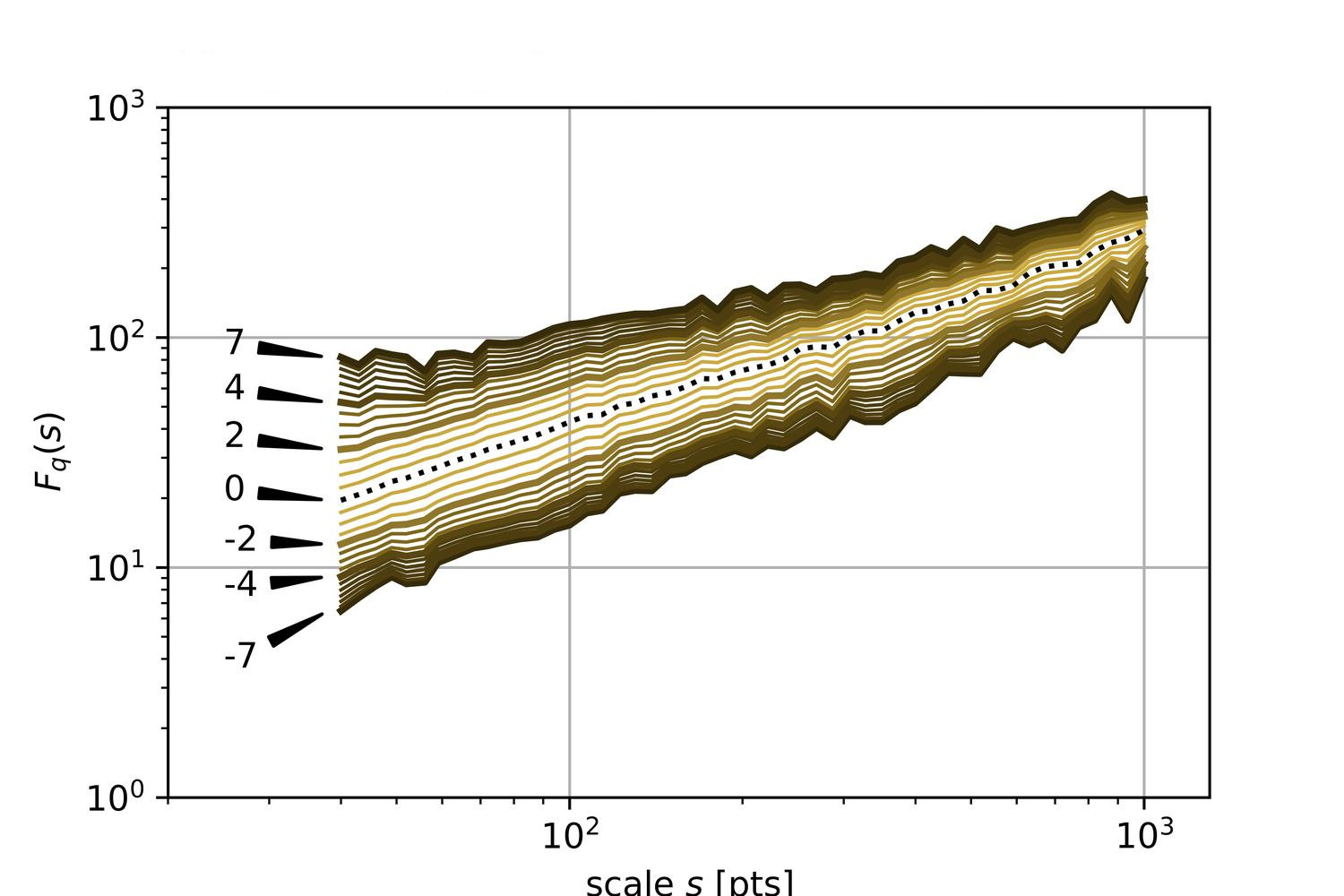}}
}
& \adjustbox{valign=t}{\includegraphics[trim={0cm 0 0 3cm}, clip, width=1.1\linewidth]{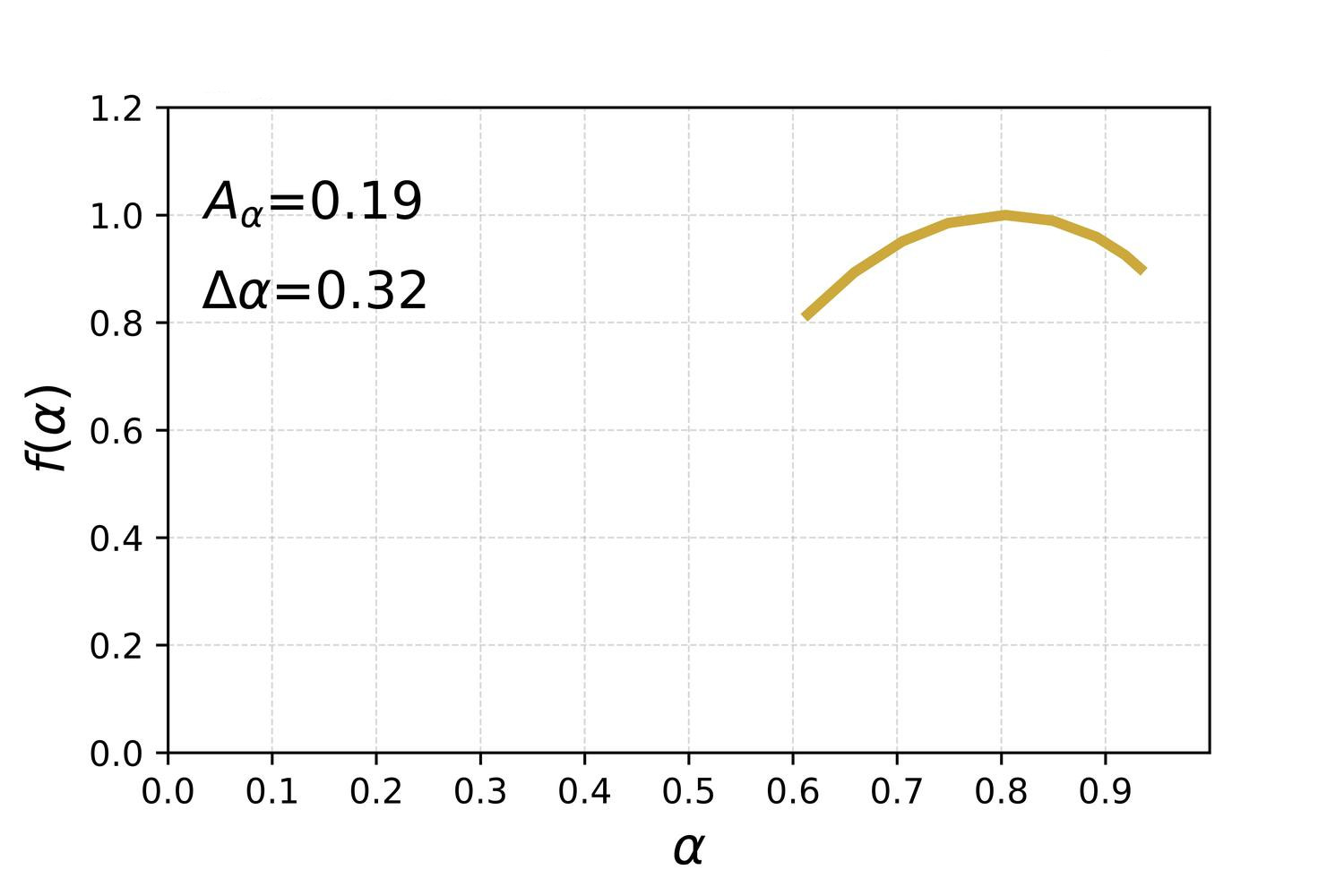}} \\
\cline{2-2}
& \adjustbox{valign=t}{\includegraphics[trim={0cm 0 0 3cm}, clip, width=1.1\linewidth]{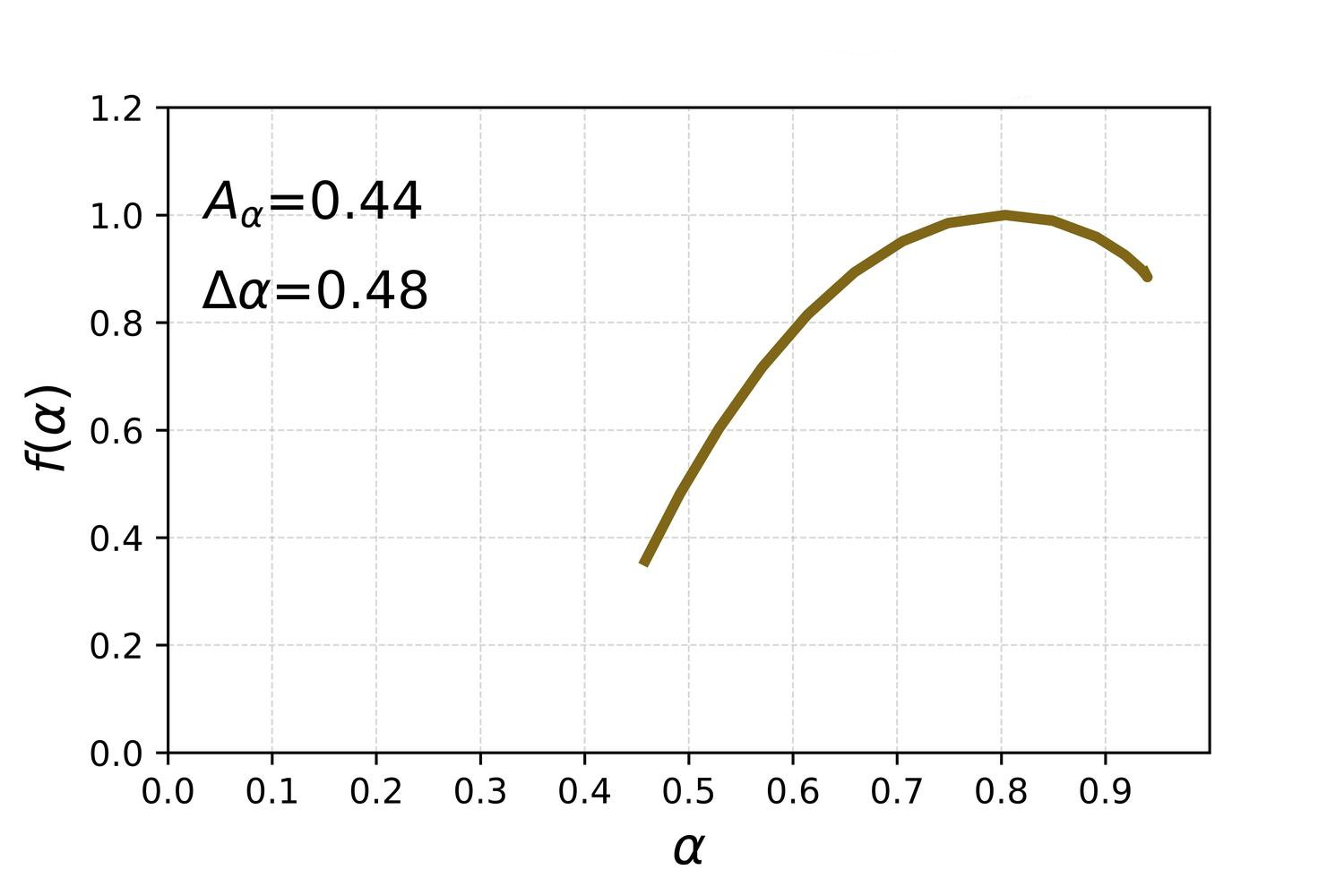}} \\
\cline{2-2}
& \adjustbox{valign=t}{\includegraphics[trim={0cm 0 0 3cm}, clip, width=1.1\linewidth]{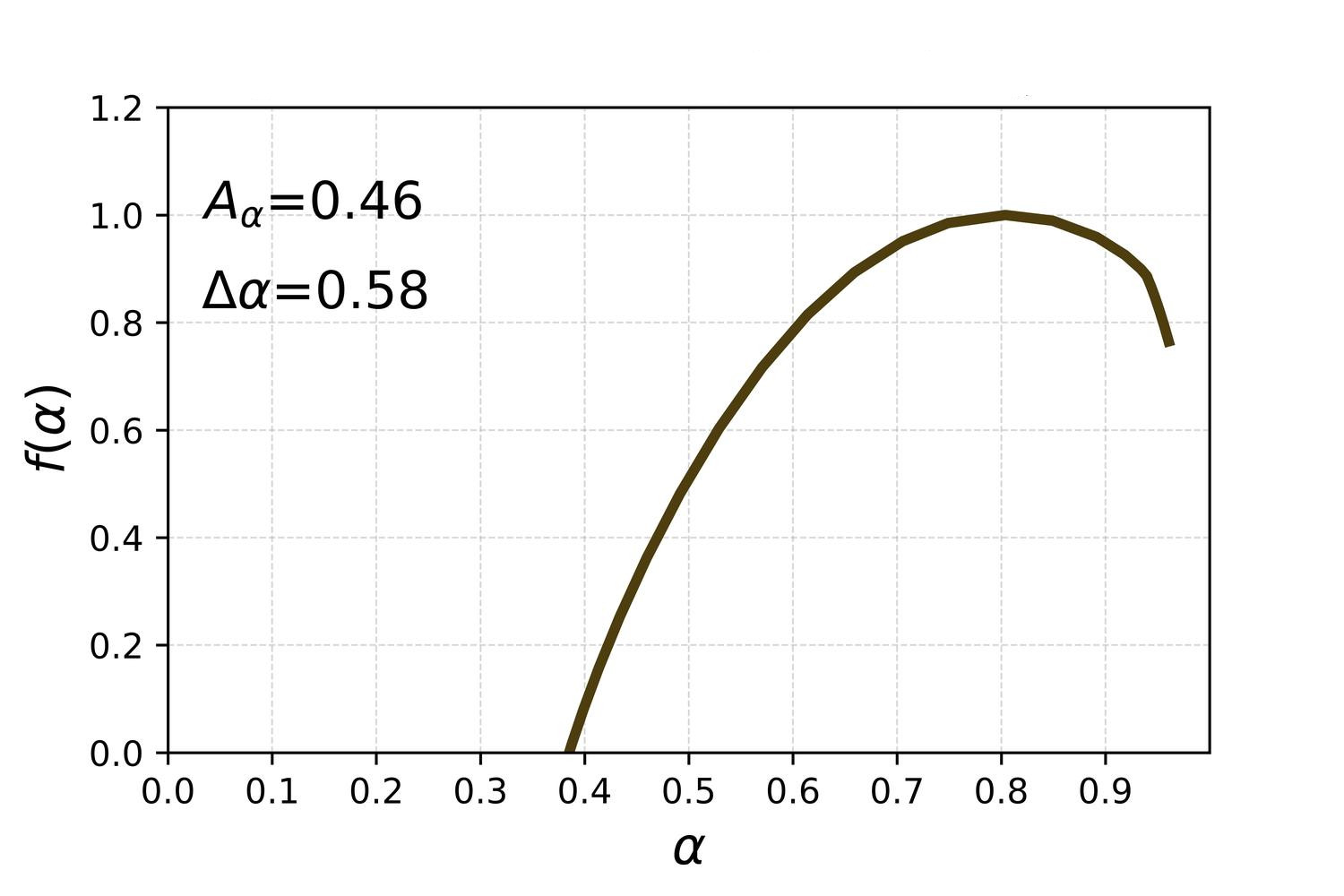}} \\
\hline
\end{tabularx}
\begin{tabularx}{\textwidth}{| >{\centering\arraybackslash}m{0.75\textwidth} | >{\centering\arraybackslash}m{0.25\textwidth} |}
\hline
\multirow{3}{*}{
\adjustbox{valign=t}{\includegraphics[trim={0cm 0 0 3cm}, clip, width=1.07\linewidth]{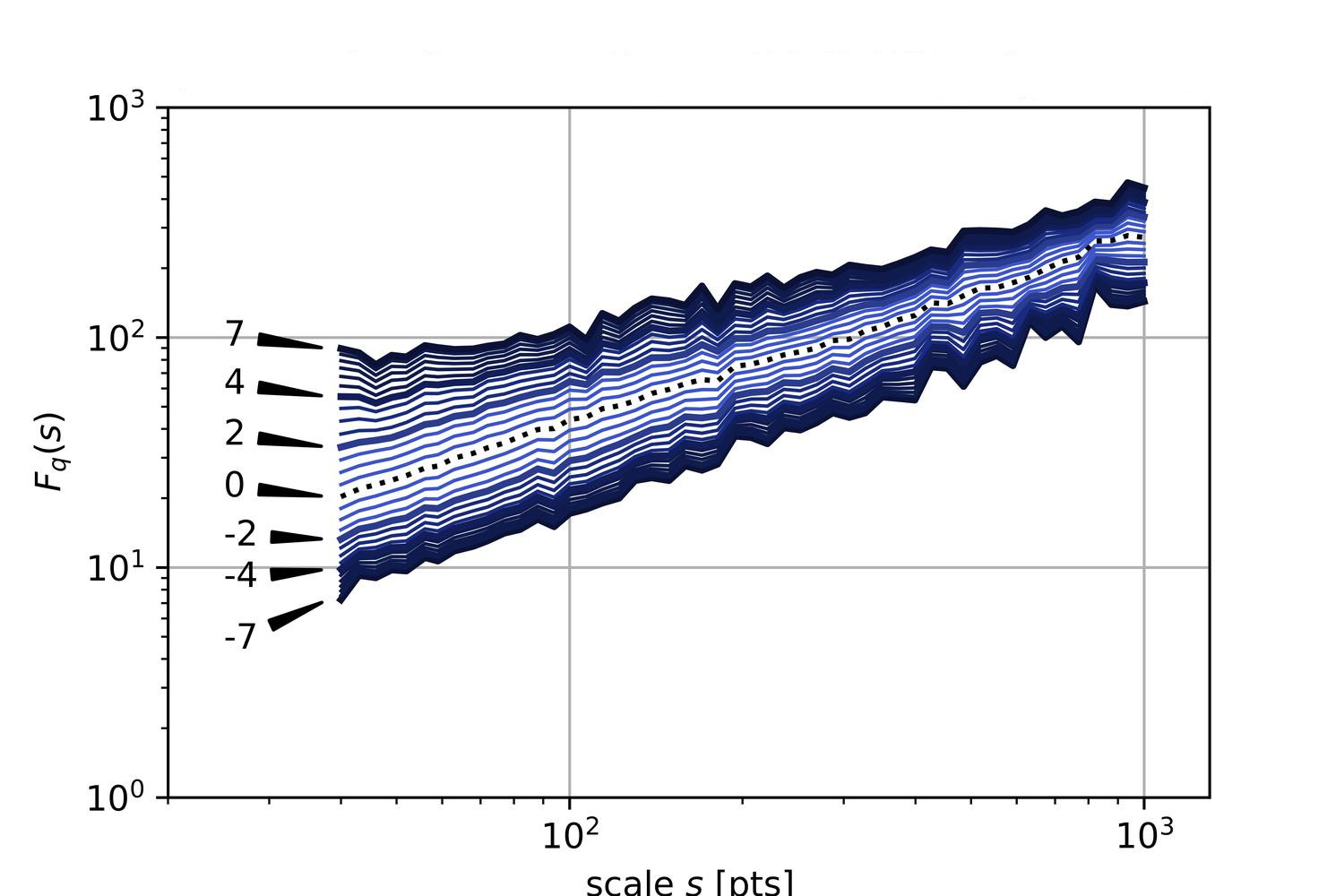}}
}
& \adjustbox{valign=t}{\includegraphics[trim={0cm 0 0 3cm}, clip, width=1.1\linewidth]{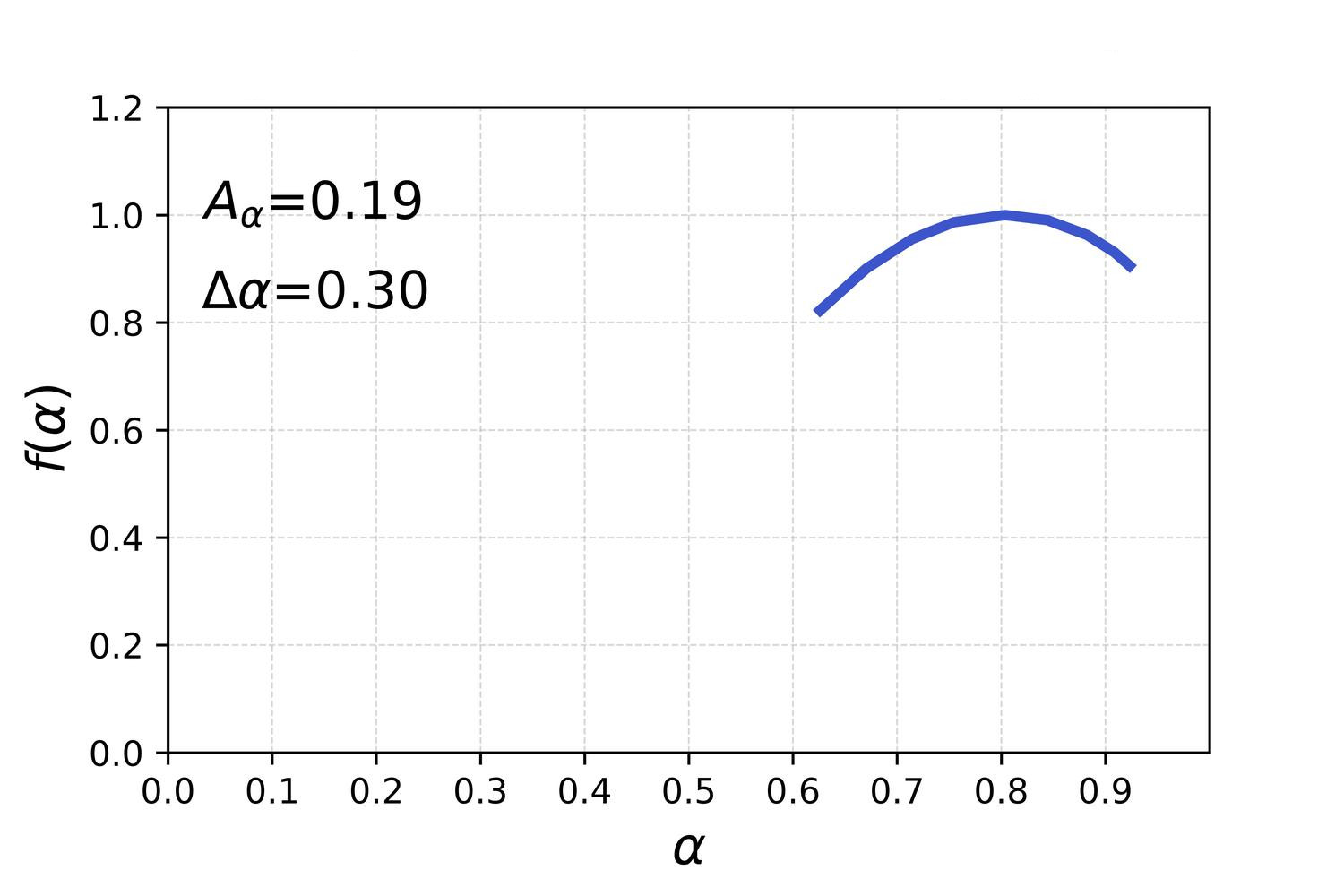}} \\
\cline{2-2}
& \adjustbox{valign=t}{\includegraphics[trim={0cm 0 0 3cm}, clip, width=1.1\linewidth]{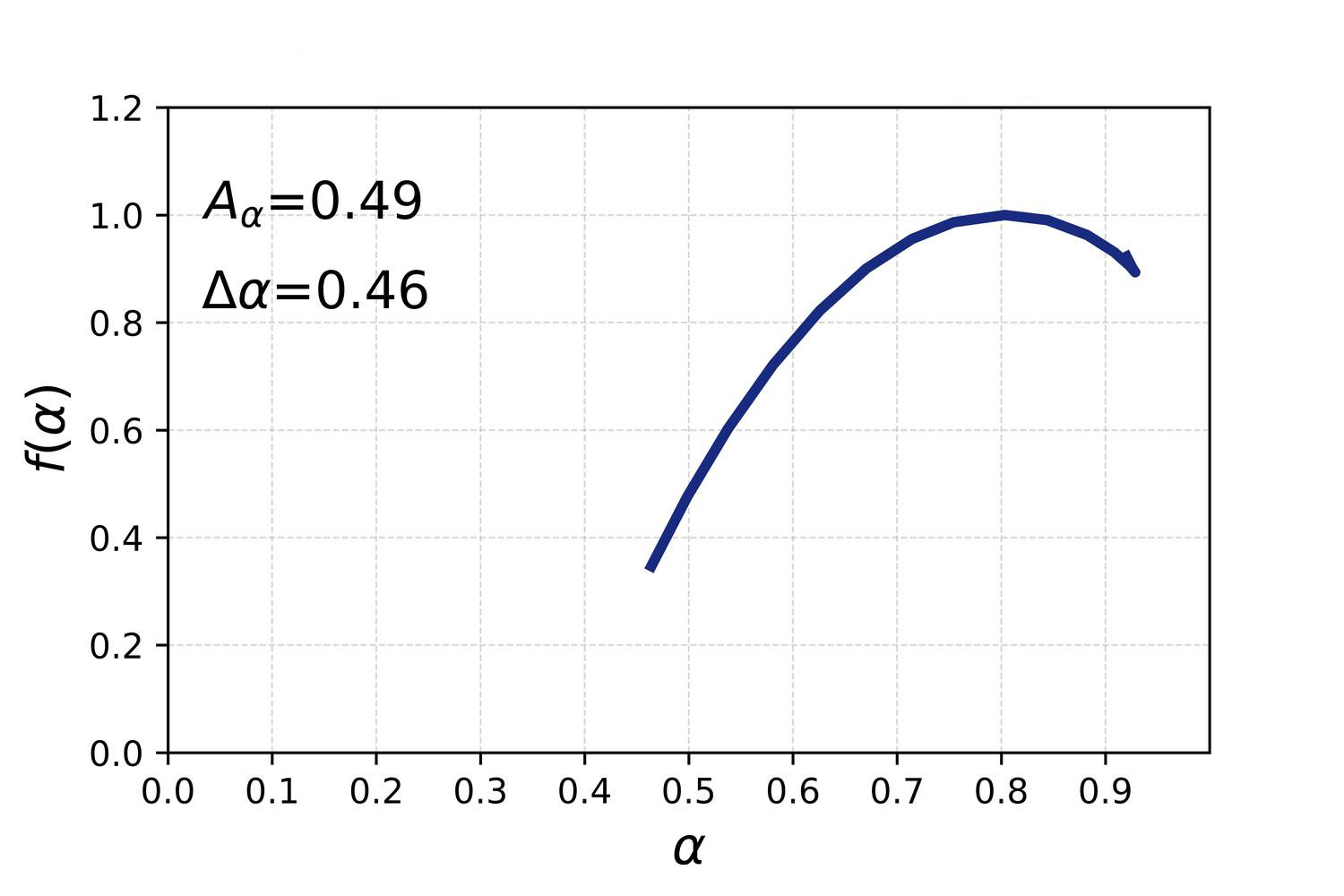}} \\
\cline{2-2}
& \adjustbox{valign=t}{\includegraphics[trim={0cm 0 0 3cm}, clip, width=1.1\linewidth]{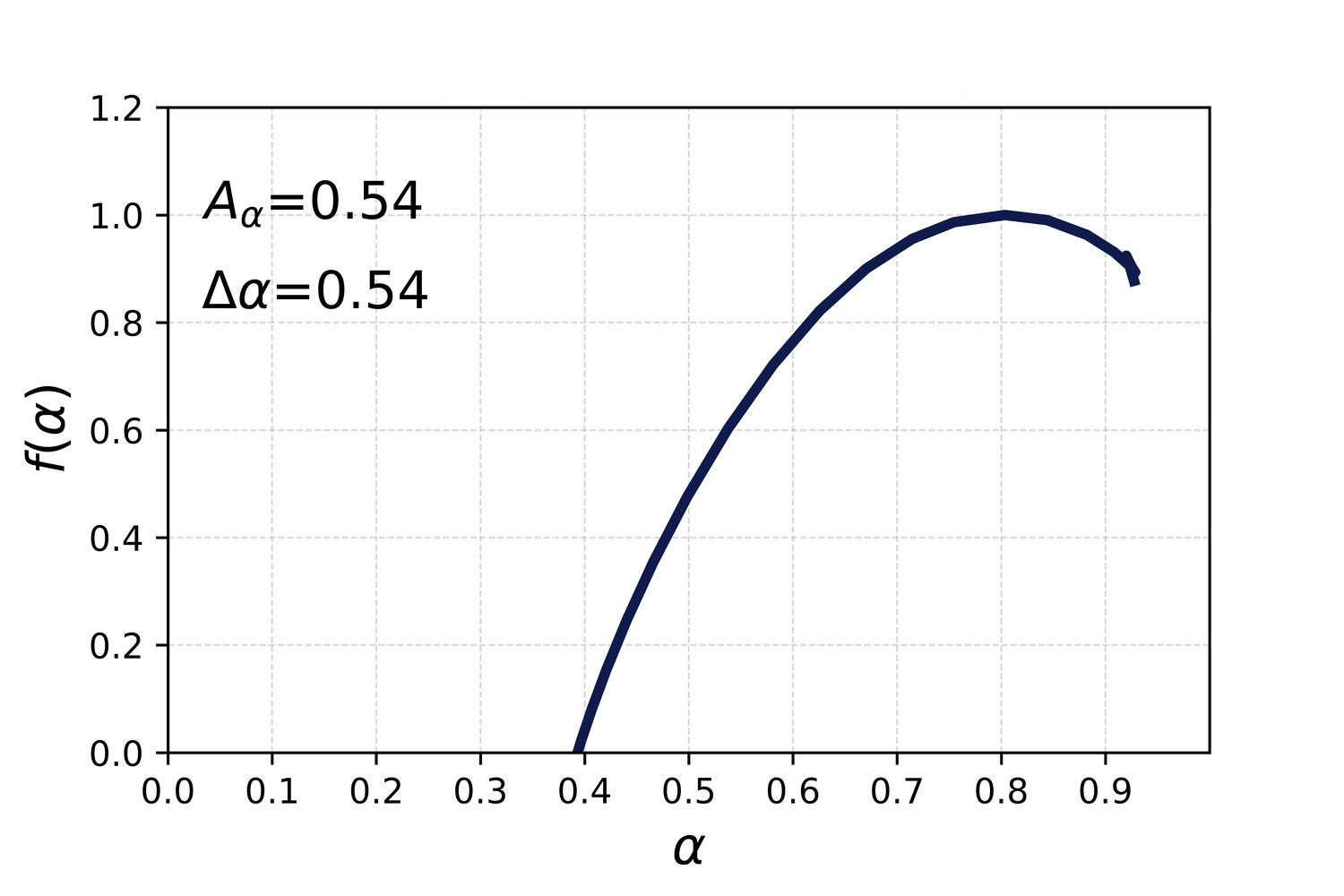}} \\
\hline
\end{tabularx}
\begin{tabularx}{\textwidth}{| >{\centering\arraybackslash}m{0.75\textwidth} | >{\centering\arraybackslash}m{0.25\textwidth} |}
\hline
\multirow{3}{*}{
\adjustbox{valign=t}{\includegraphics[trim={0cm 0 0 3cm}, clip, width=1.07\linewidth]{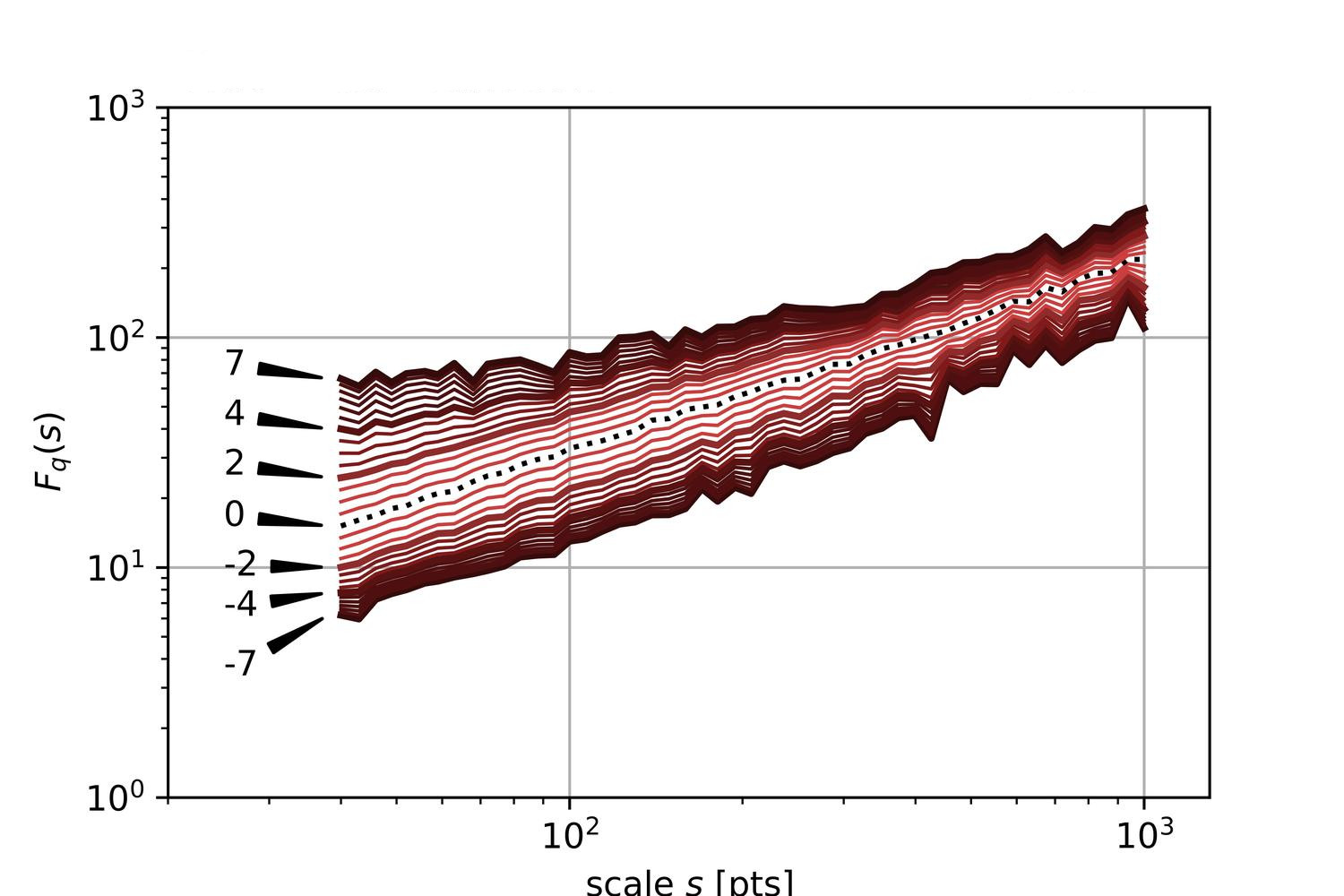}}
}
& \adjustbox{valign=t}{\includegraphics[trim={0cm 0 0 3cm}, clip, width=1.1\linewidth]{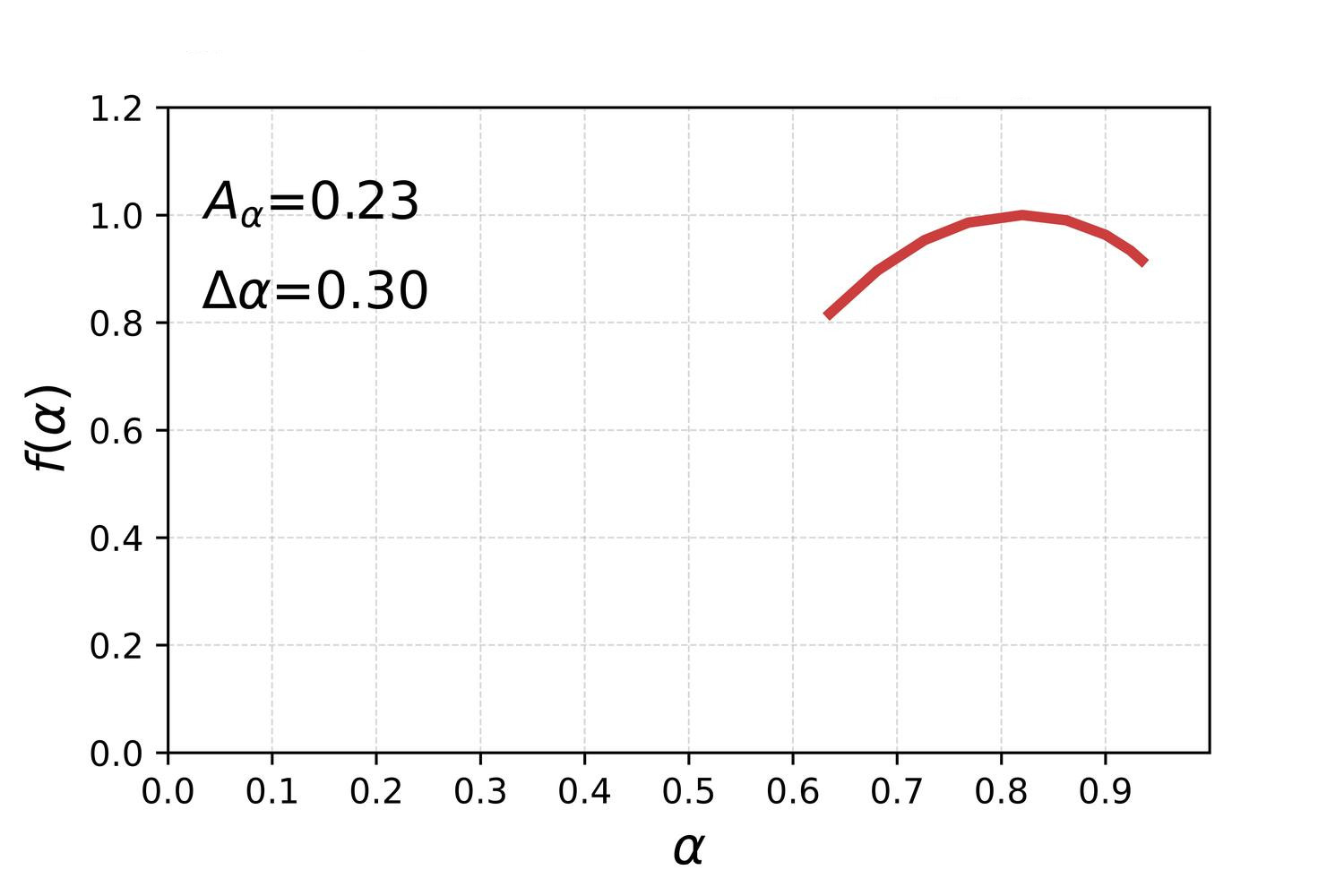}} \\
\cline{2-2}
& \adjustbox{valign=t}{\includegraphics[trim={0cm 0 0 3cm}, clip, width=1.1\linewidth]{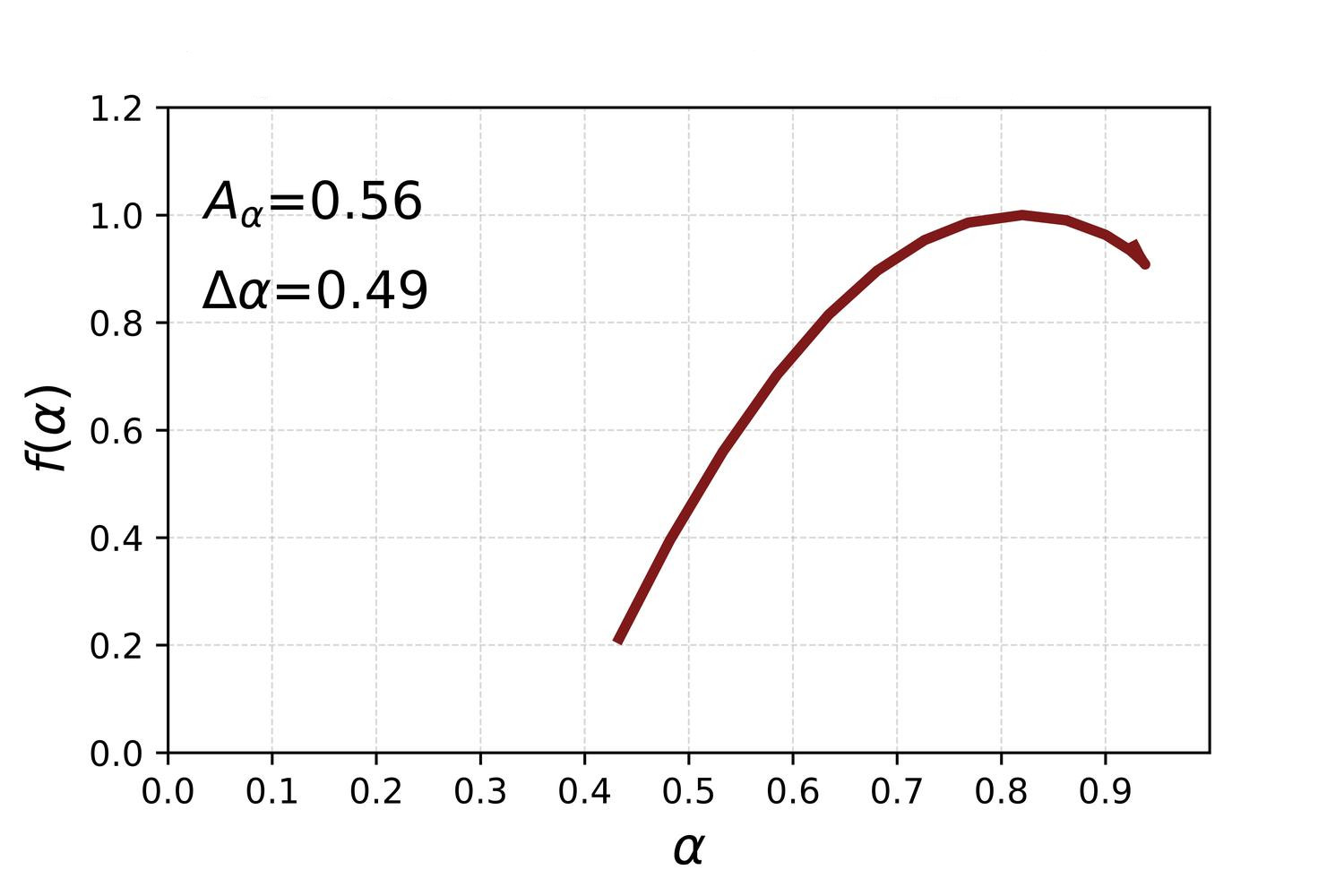}} \\
\cline{2-2}
& \adjustbox{valign=t}{\includegraphics[trim={0cm 0 0 3cm}, clip, width=1.1\linewidth]{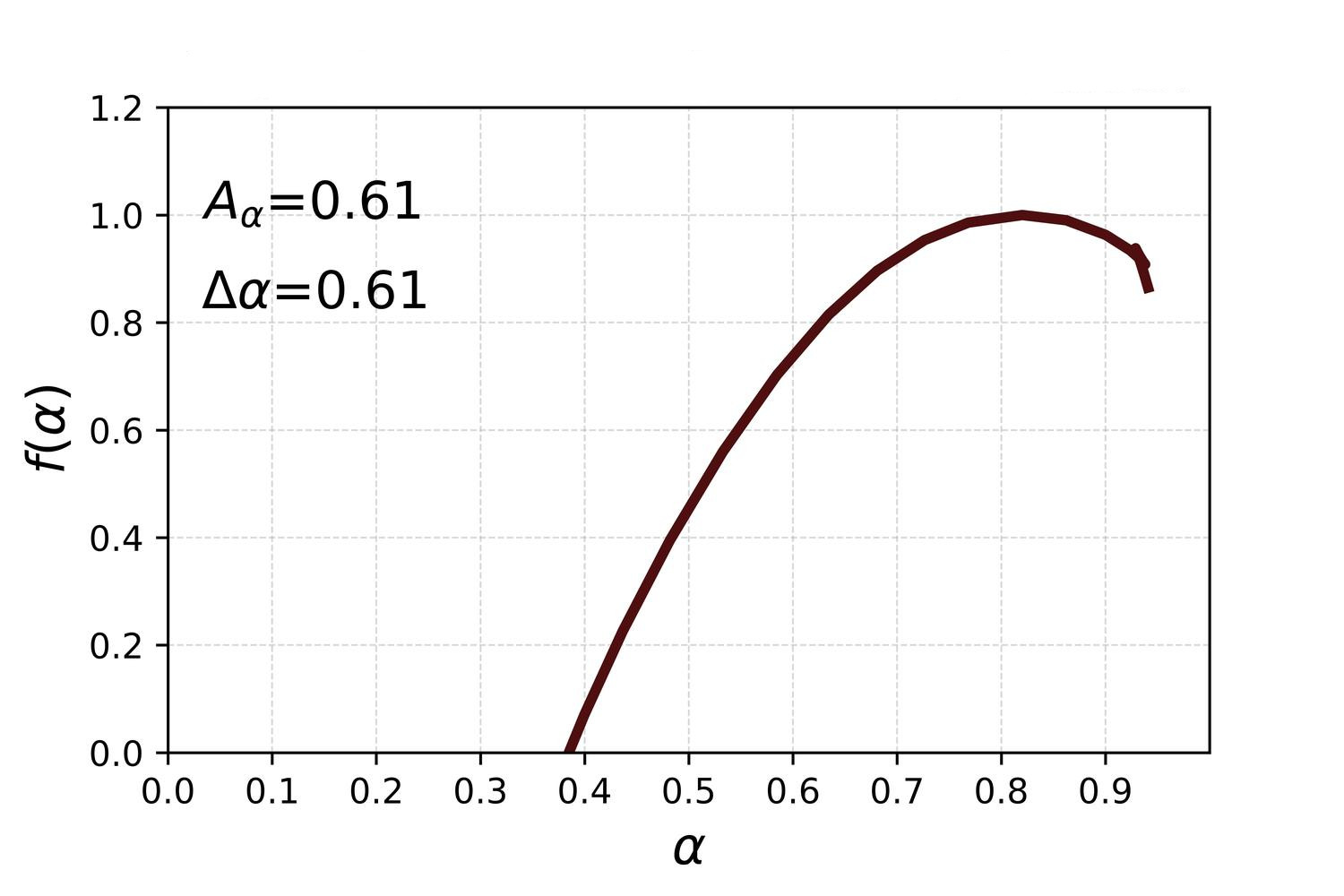}} \\
\hline
\end{tabularx}
\end{adjustwidth}
\caption{(Main) $q$-dependent fluctuation functions $F_q(s)$ calculated for time series of SLV in the Spanish original text of \textit{Hopscotch} (\textbf{top}) and its two translations into English (\textbf{middle}) and Polish (\textbf{bottom}). The printed order of chapters is considered. The plots of $F_q(s)$ for particular values of $q$ are indicated by arrows. (Side) Singularity spectra $f(\alpha)$ associated with the exponents $h(q)$ calculated from the scaling regions of the respective functions $F_q(s)$ for three intervals: $q \in \left[ -2, 2\right]$ (\textbf{top right}), $q \in \left[ -4, 4\right]$ (\textbf{middle right}), and $q \in \left[ -7, 7\right]$ (\textbf{bottom right}). In each case the asymmetry coefficient $A_{\alpha}$ and the singularity spectrum width $\Delta \alpha$ are also given.}
\label{fig::slv.Fq.printed}
\end{figure}

\begin{figure}[H]
\begin{adjustwidth}{-\extralength}{0cm}
\centering
\setlength\tabcolsep{1pt}
\begin{tabularx}{\textwidth}{| >{\centering\arraybackslash}m{0.75\textwidth} | >{\centering\arraybackslash}m{0.25\textwidth} |}
\hline
\multirow{3}{*}{
\adjustbox{valign=t}{\includegraphics[trim={0cm 0 0 3cm}, clip, width=1.07\linewidth]{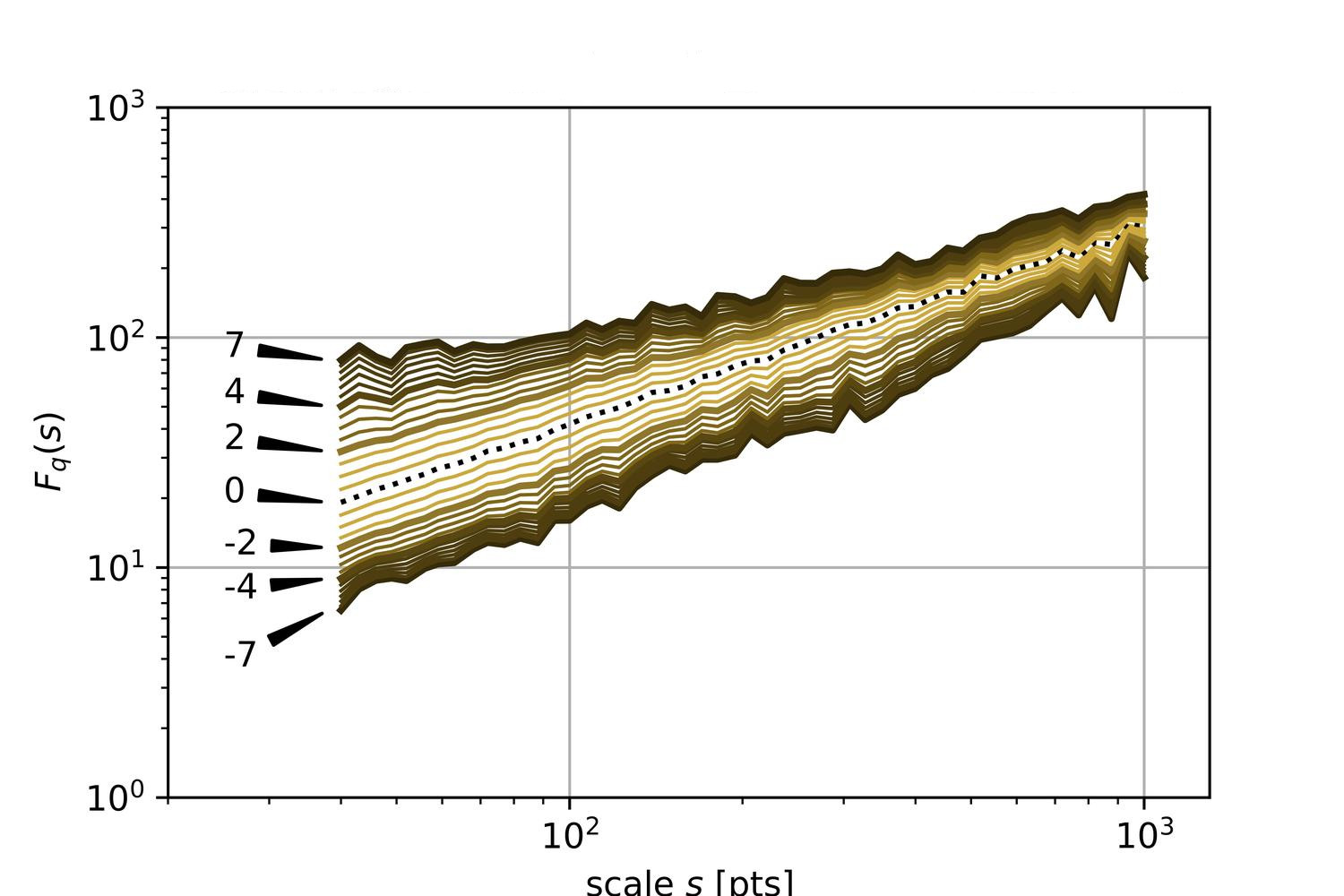}}
}
& \adjustbox{valign=t}{\includegraphics[trim={0cm 0 0 3cm}, clip, width=1.1\linewidth]{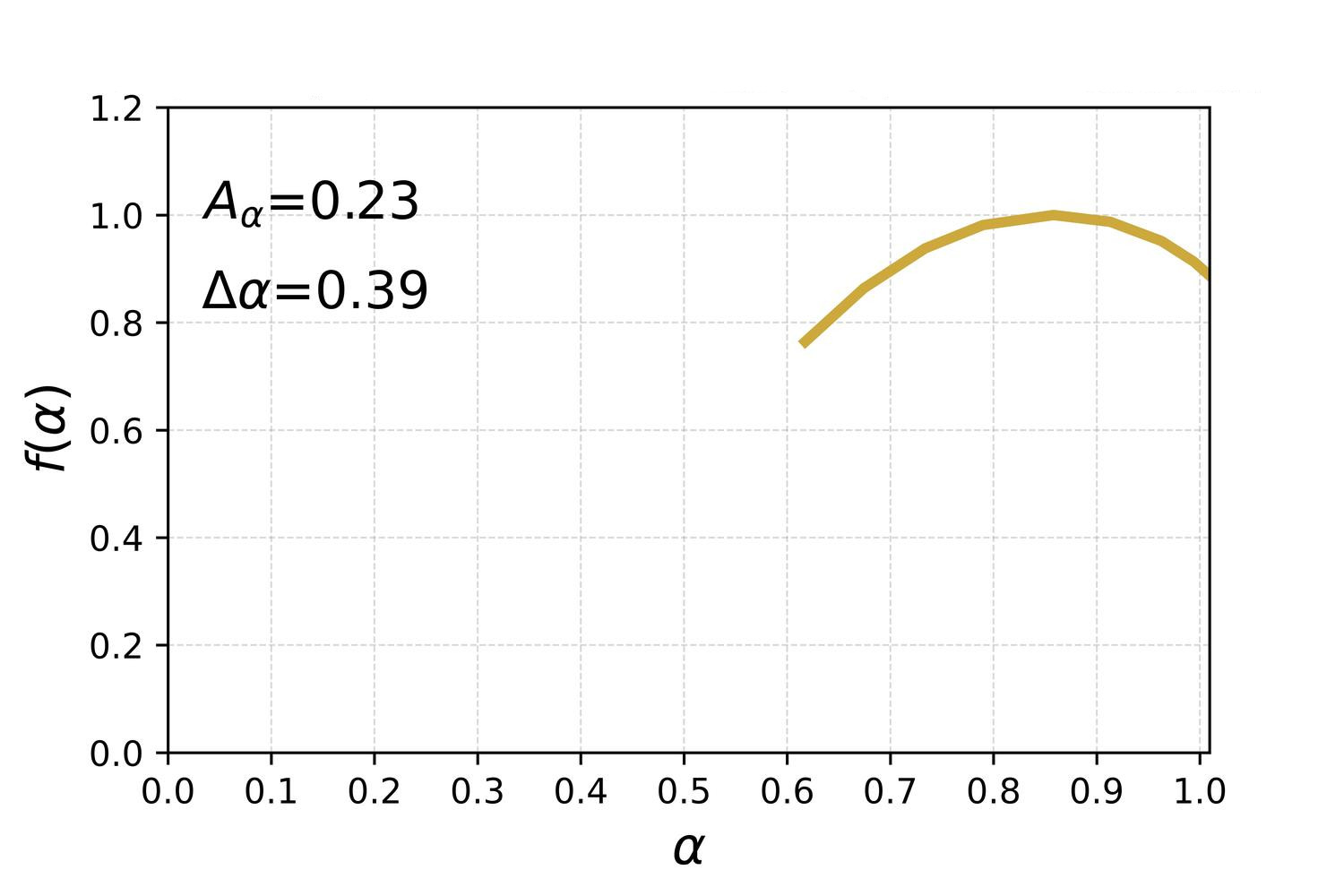}} \\
\cline{2-2}
& \adjustbox{valign=t}{\includegraphics[trim={0cm 0 0 3cm}, clip, width=1.1\linewidth]{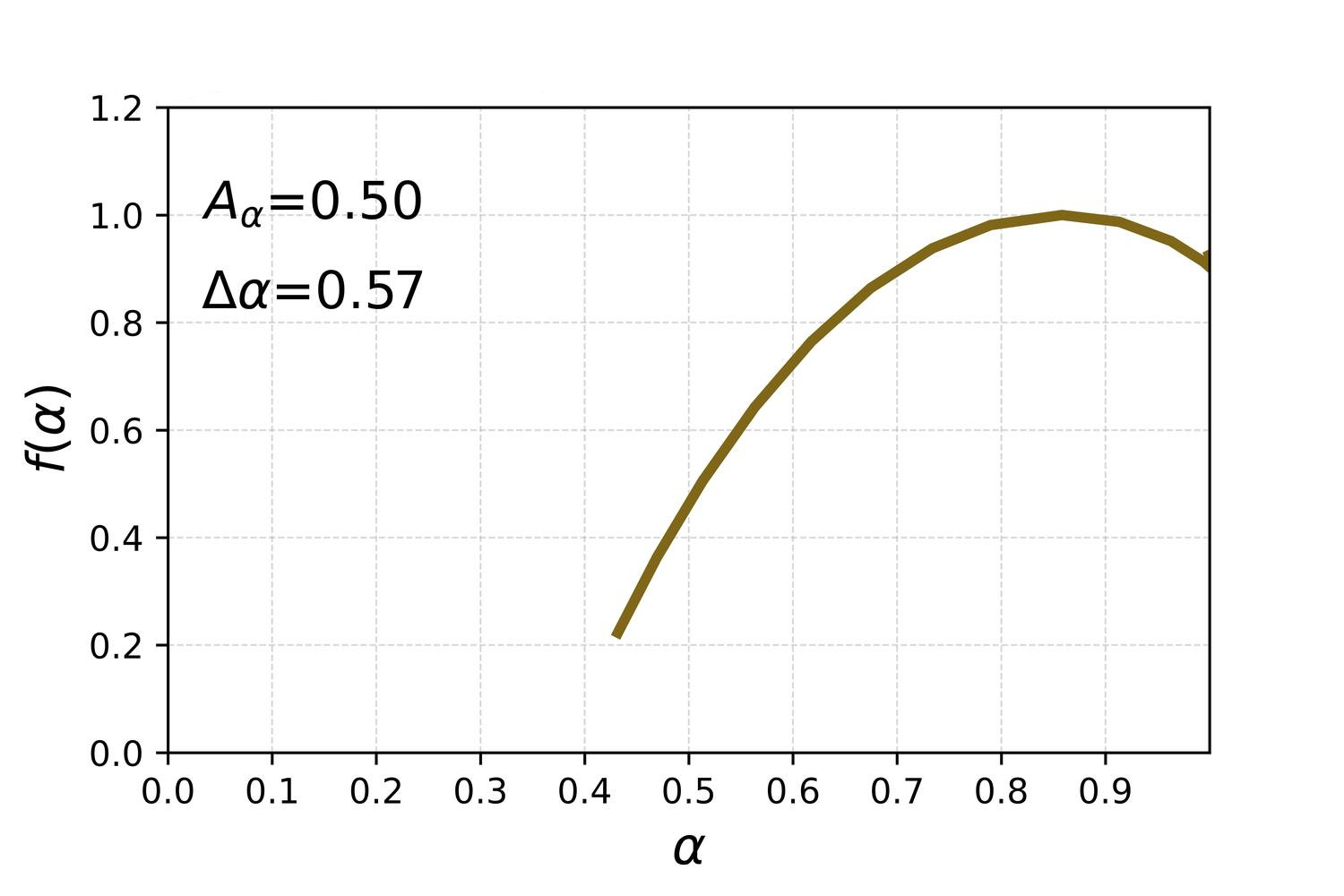}} \\
\cline{2-2}
& \adjustbox{valign=t}{\includegraphics[trim={0cm 0 0 3cm}, clip, width=1.1\linewidth]{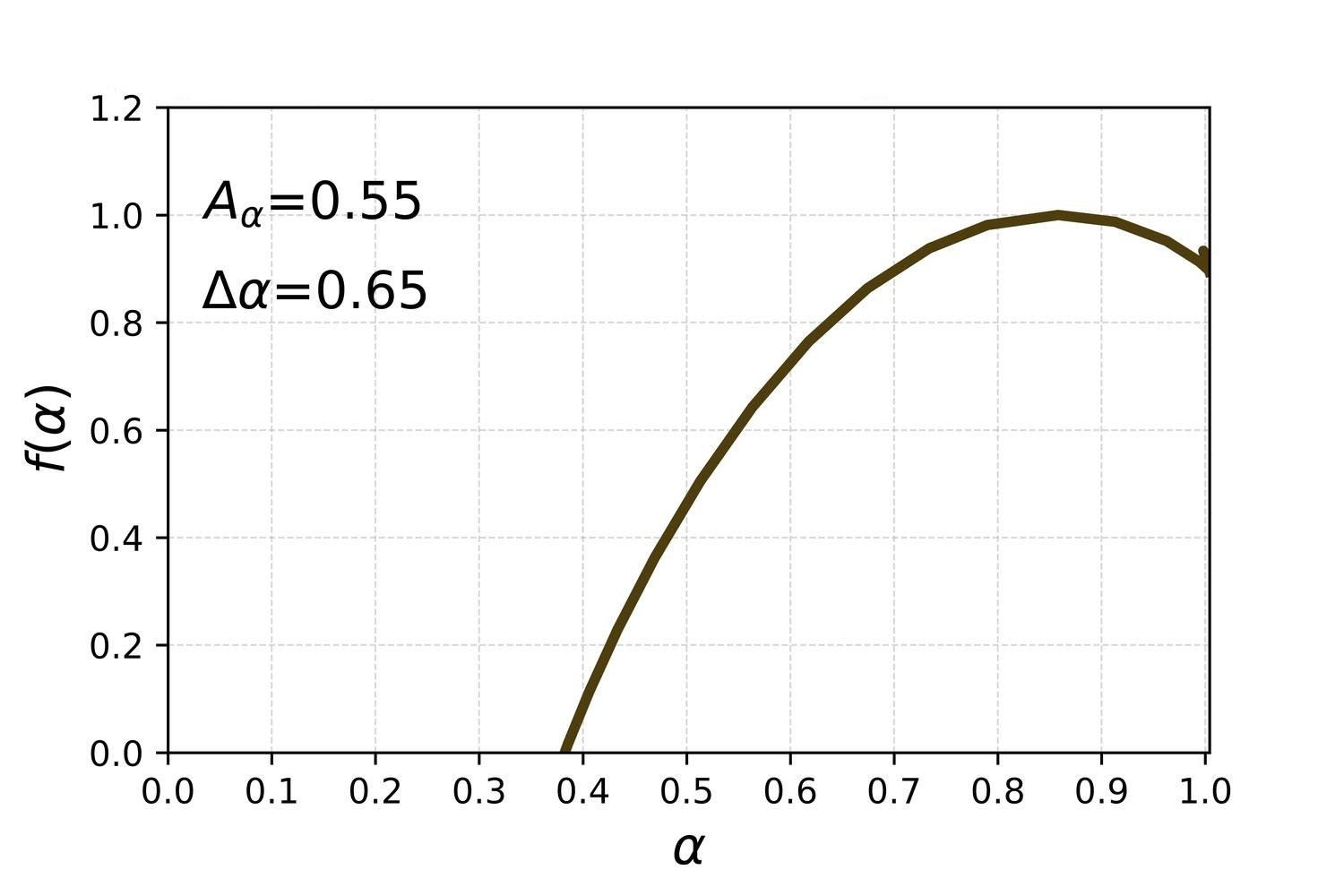}} \\
\hline
\end{tabularx}
\begin{tabularx}{\textwidth}{| >{\centering\arraybackslash}m{0.75\textwidth} | >{\centering\arraybackslash}m{0.25\textwidth} |}
\hline
\multirow{3}{*}{
\adjustbox{valign=t}{\includegraphics[trim={0cm 0 0 3cm}, clip, width=1.07\linewidth]{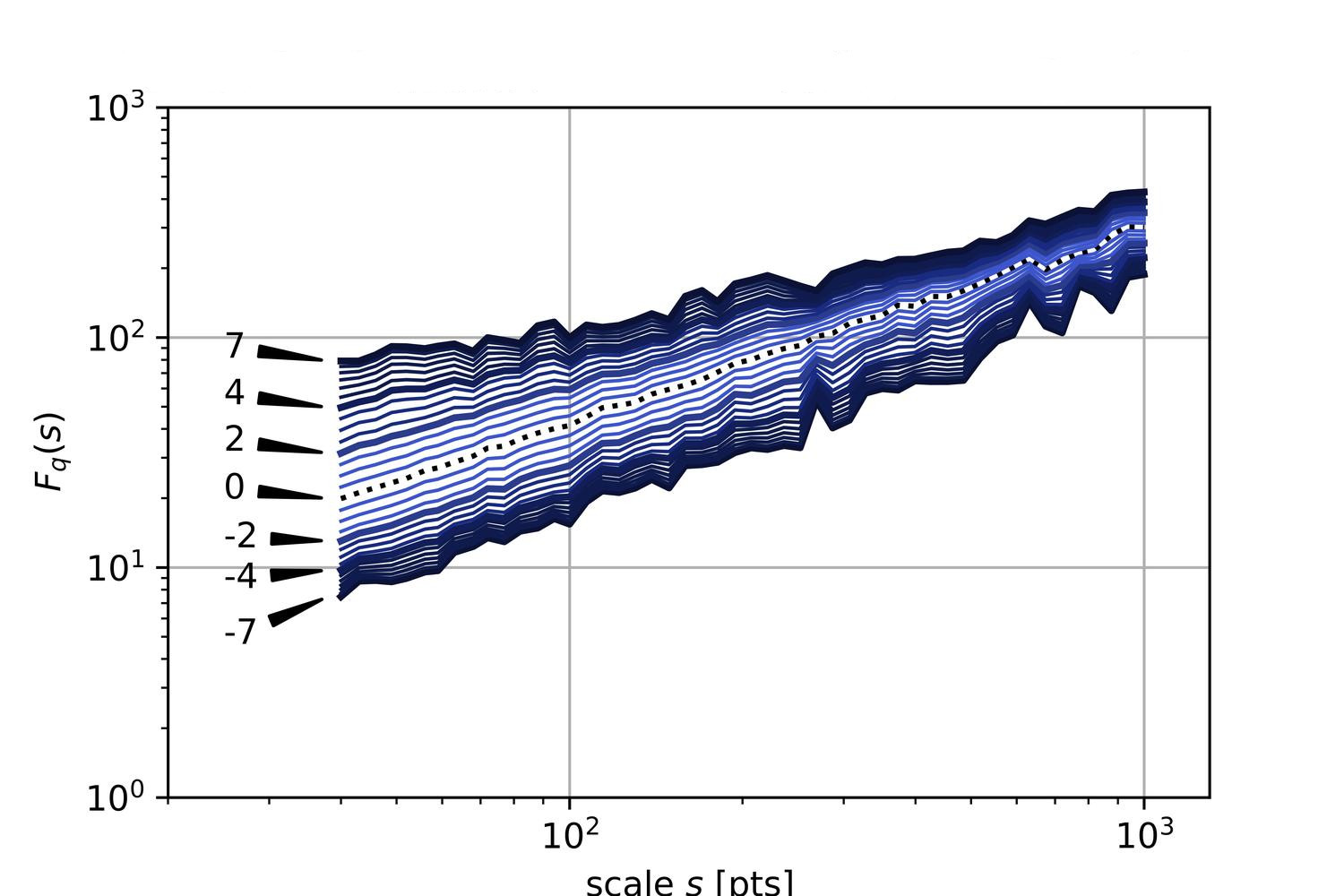}}
}
& \adjustbox{valign=t}{\includegraphics[trim={0cm 0 0 3cm}, clip, width=1.1\linewidth]{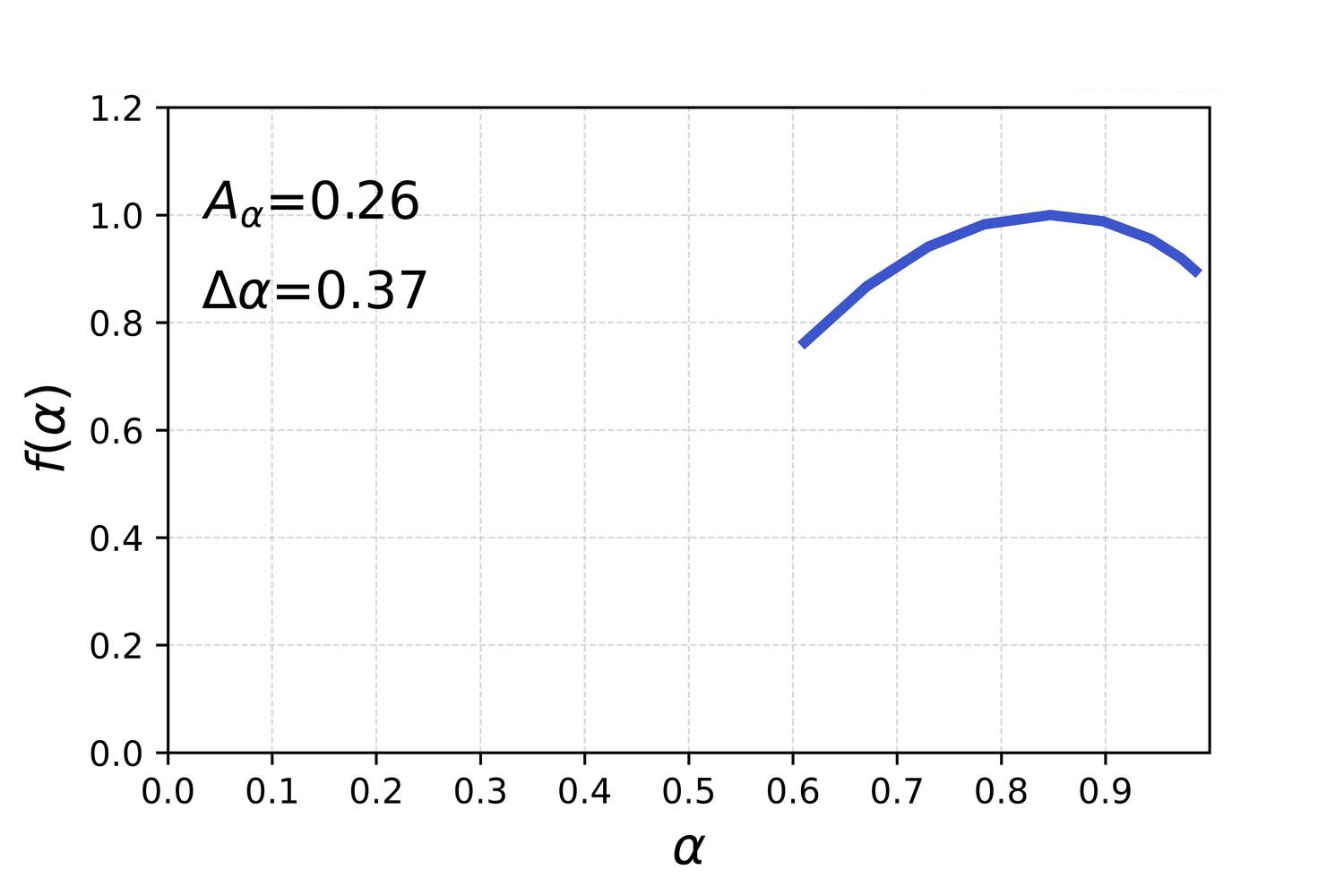}} \\
\cline{2-2}
& \adjustbox{valign=t}{\includegraphics[trim={0cm 0 0 3cm}, clip, width=1.1\linewidth]{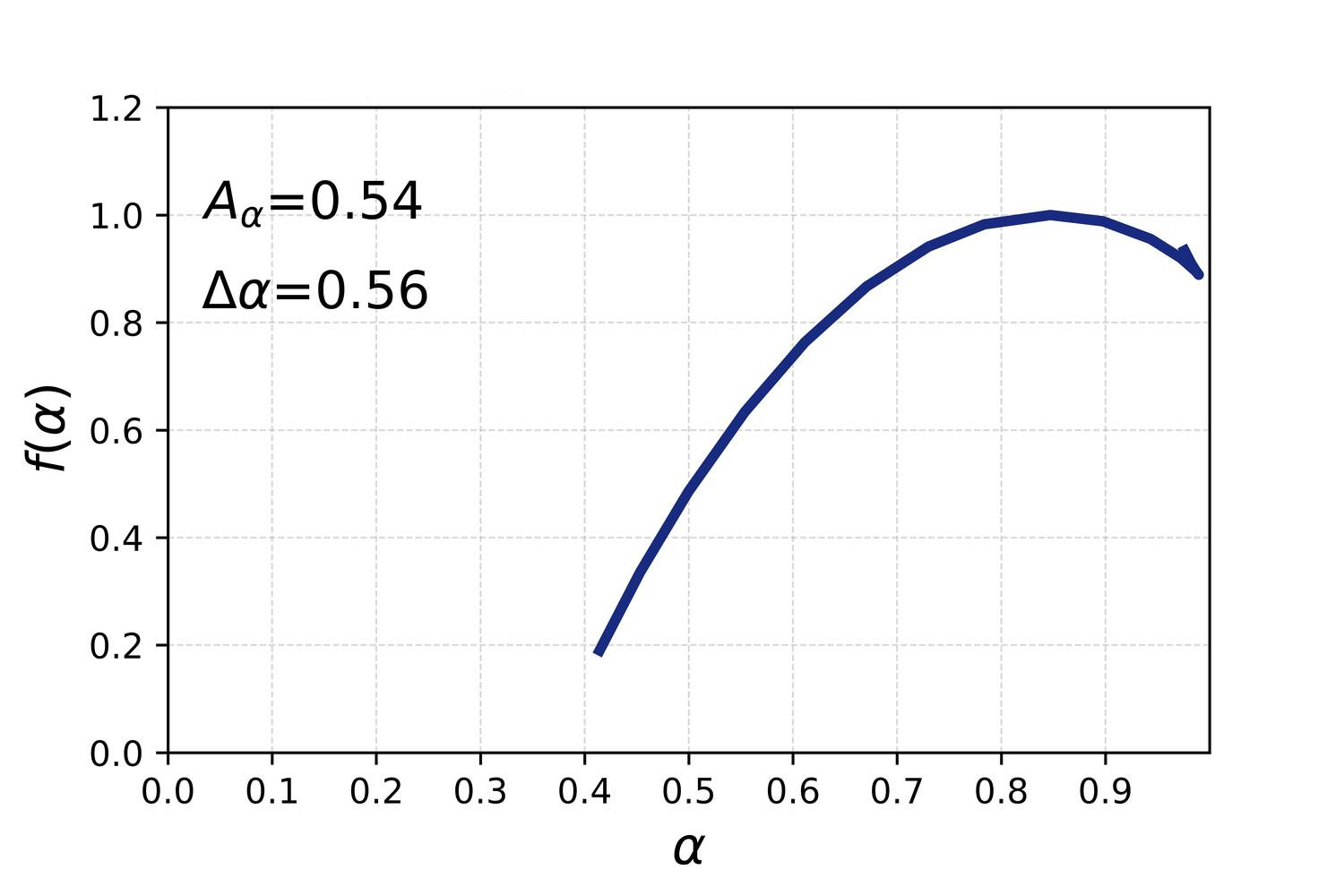}} \\
\cline{2-2}
& \adjustbox{valign=t}{\includegraphics[trim={0cm 0 0 3cm}, clip, width=1.1\linewidth]{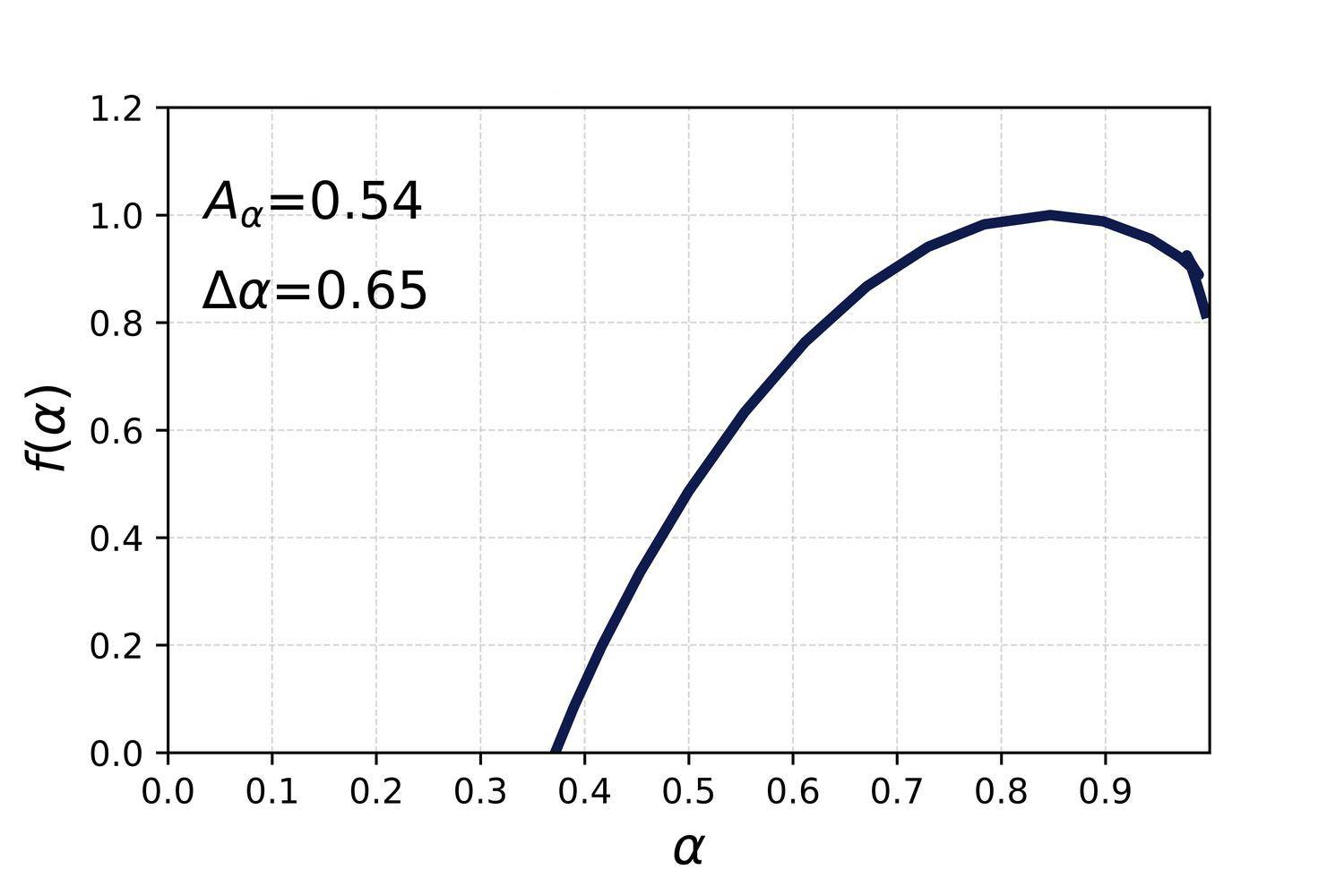}} \\
\hline
\end{tabularx}
\begin{tabularx}{\textwidth}{| >{\centering\arraybackslash}m{0.75\textwidth} | >{\centering\arraybackslash}m{0.25\textwidth} |}
\hline
\multirow{3}{*}{
\adjustbox{valign=t}{\includegraphics[trim={0cm 0 0 3cm}, clip, width=1.07\linewidth]{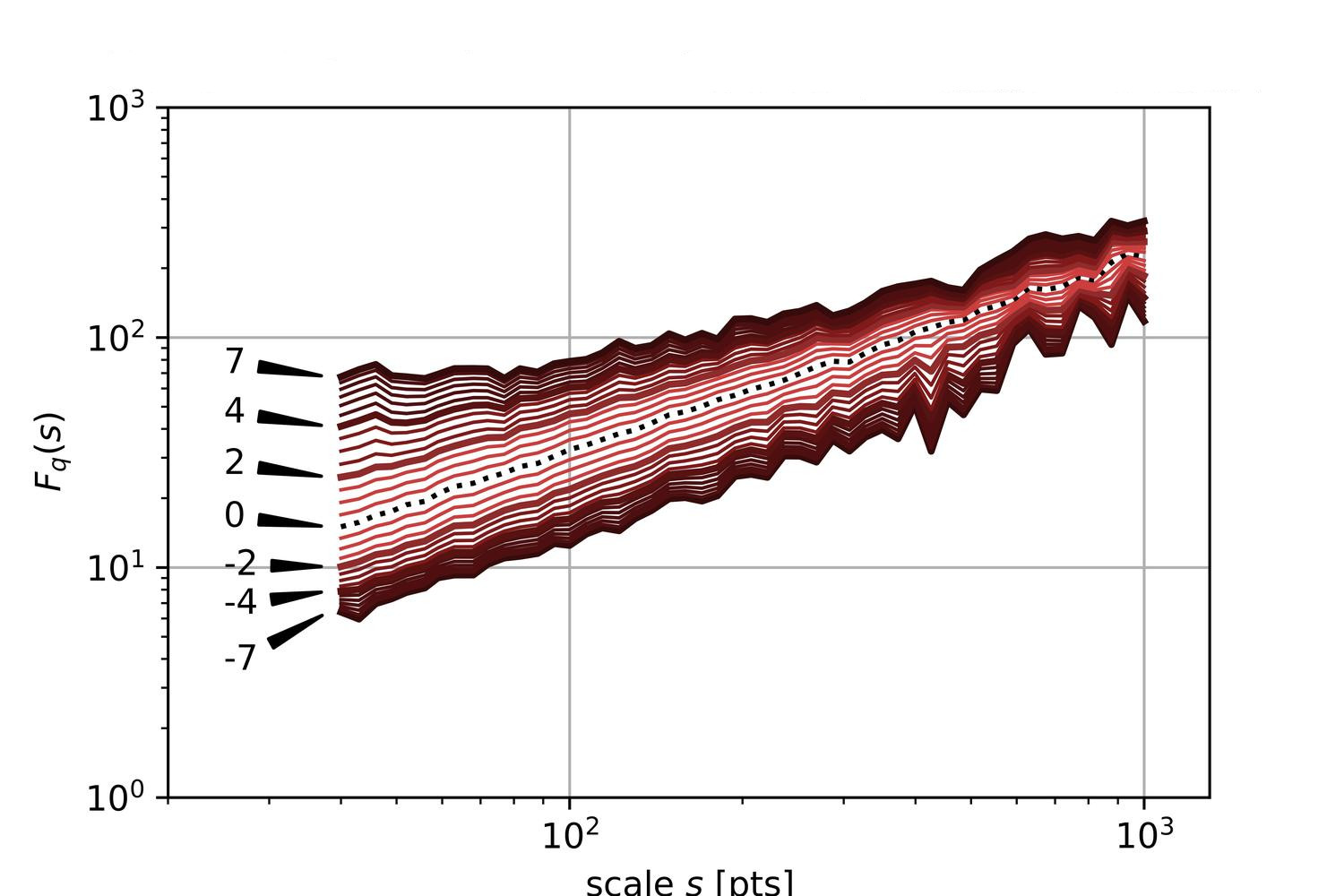}}
}
& \adjustbox{valign=t}{\includegraphics[trim={0cm 0 0 3cm}, clip, width=1.1\linewidth]{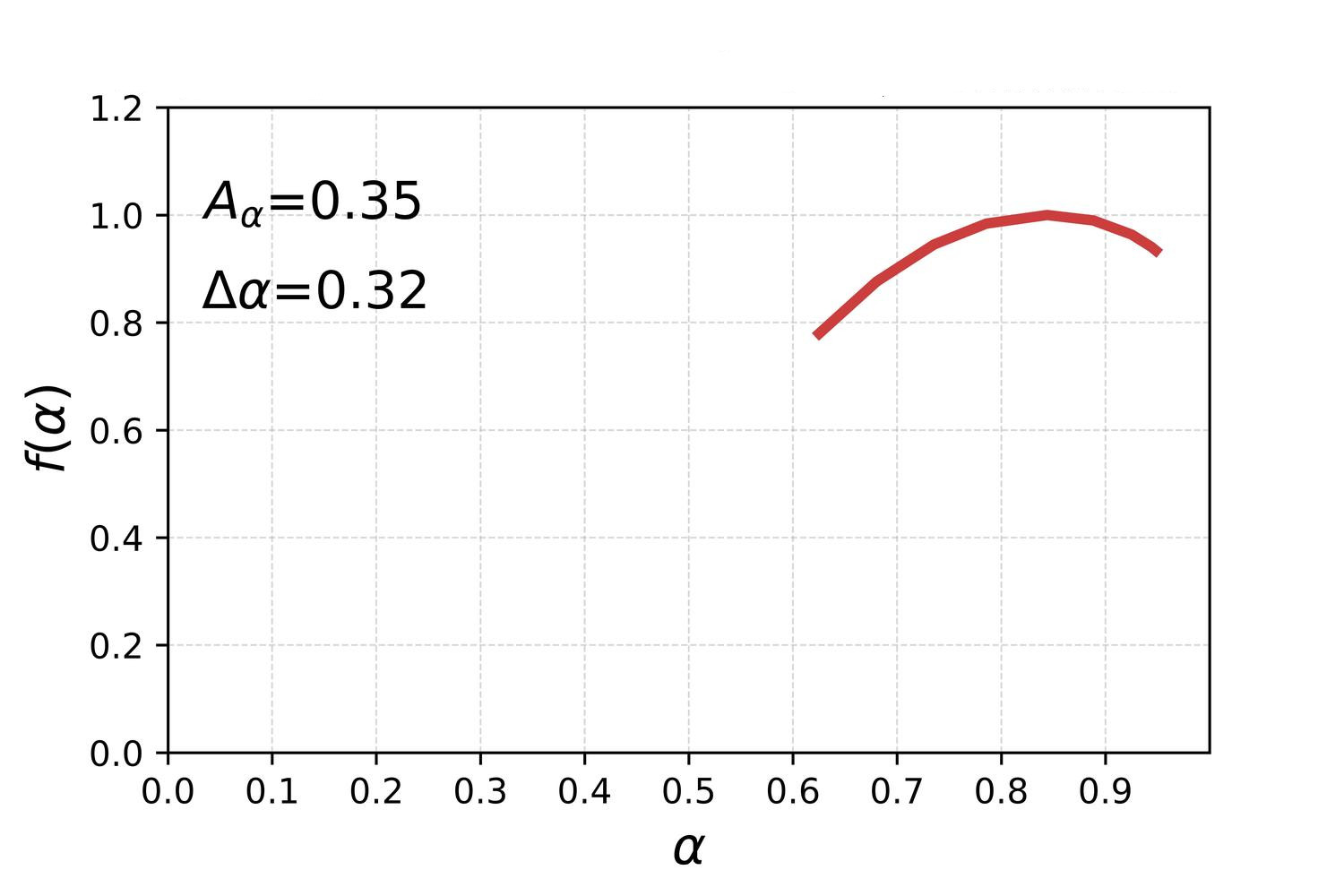}} \\
\cline{2-2}
& \adjustbox{valign=t}{\includegraphics[trim={0cm 0 0 3cm}, clip, width=1.1\linewidth]{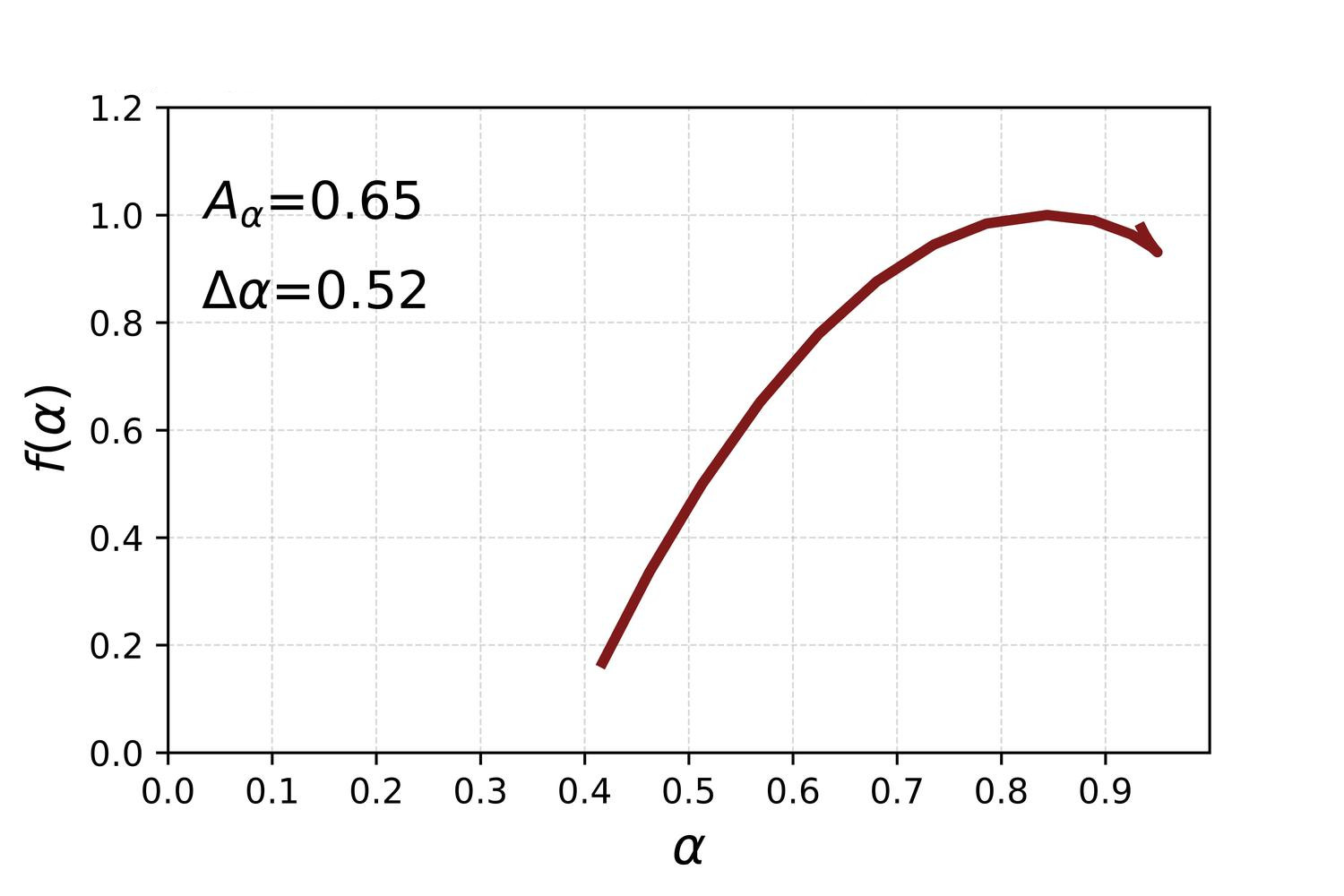}} \\
\cline{2-2}
& \adjustbox{valign=t}{\includegraphics[trim={0cm 0 0 3cm}, clip, width=1.1\linewidth]{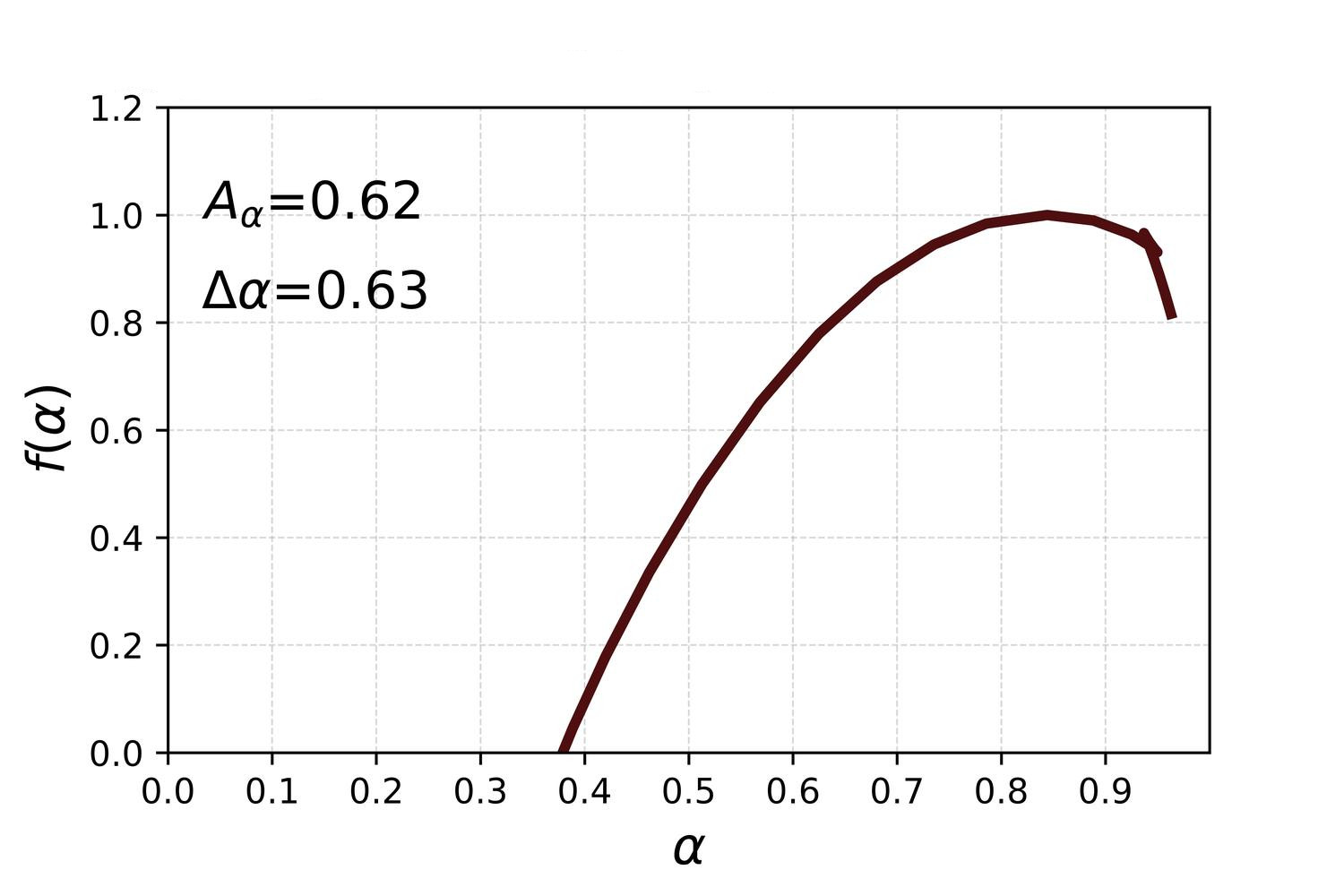}} \\
\hline
\end{tabularx}
\end{adjustwidth}     
\caption{The same functions as in Figure~\ref{fig::slv.Fq.printed} for the recommended order of chapters.}
\label{fig::slv.Fq.recommended}
\end{figure}

\begin{figure}[H]
\begin{adjustwidth}{-\extralength}{0cm}
\centering
\setlength\tabcolsep{1pt}
\begin{tabularx}{\textwidth}{| >{\centering\arraybackslash}m{0.75\textwidth} | >{\centering\arraybackslash}m{0.25\textwidth} |}
\hline
\multirow{3}{*}{
\adjustbox{valign=t}{\includegraphics[trim={0cm 0 0 3cm}, clip, width=1.07\linewidth]{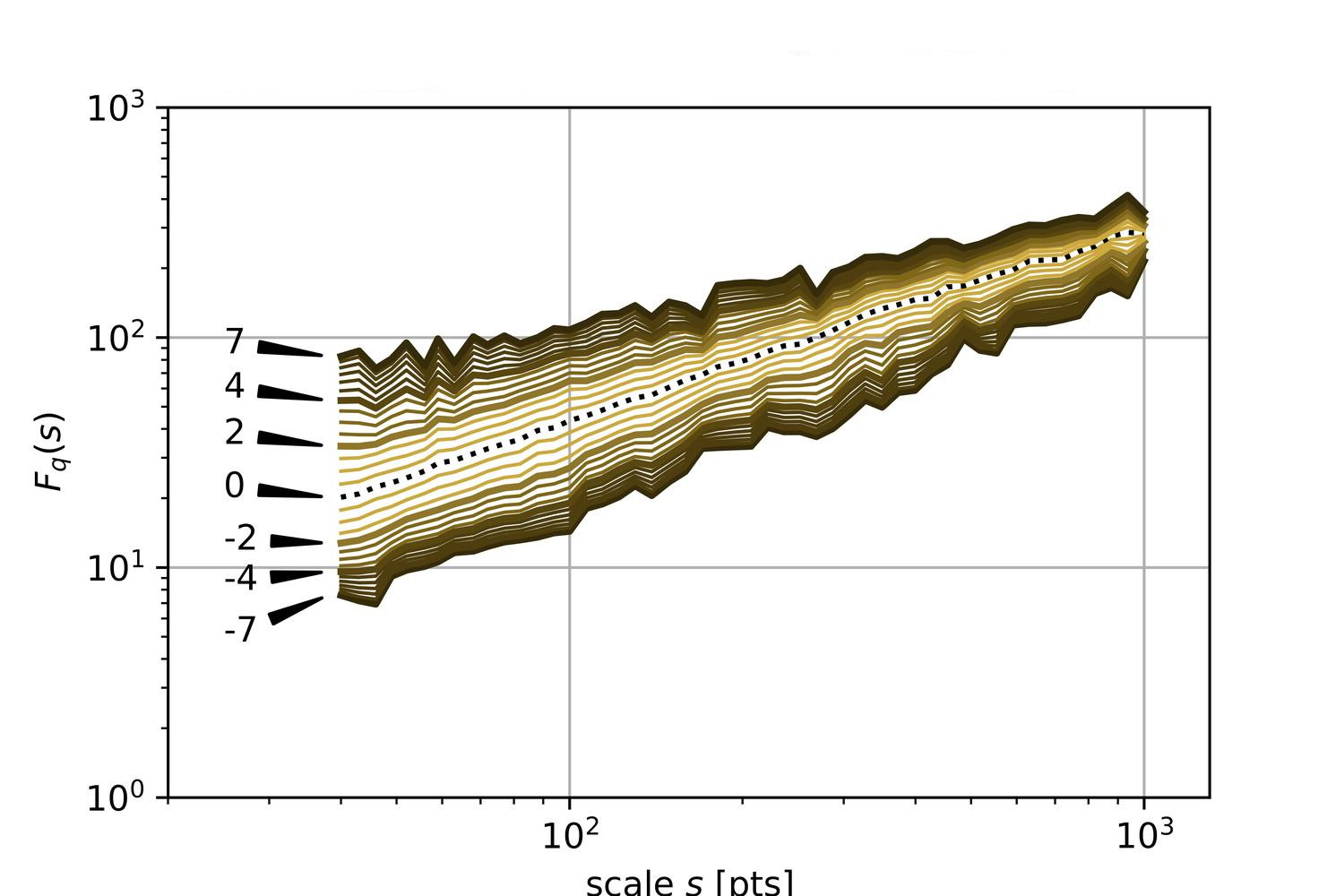}}
}
& \adjustbox{valign=t}{\includegraphics[trim={0cm 0 0 3cm}, clip, width=1.1\linewidth]{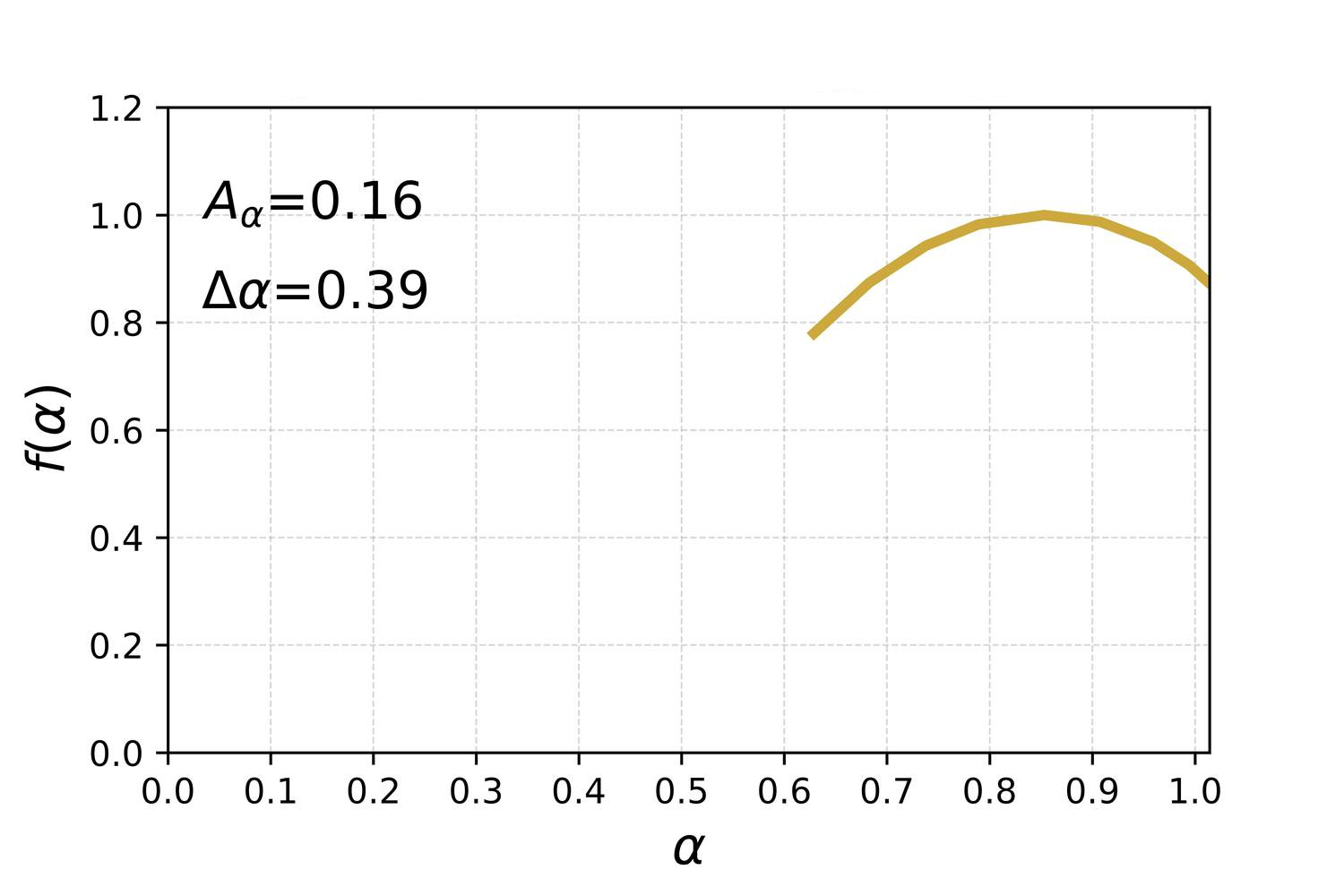}} \\
\cline{2-2}
& \adjustbox{valign=t}{\includegraphics[trim={0cm 0 0 3cm}, clip, width=1.1\linewidth]{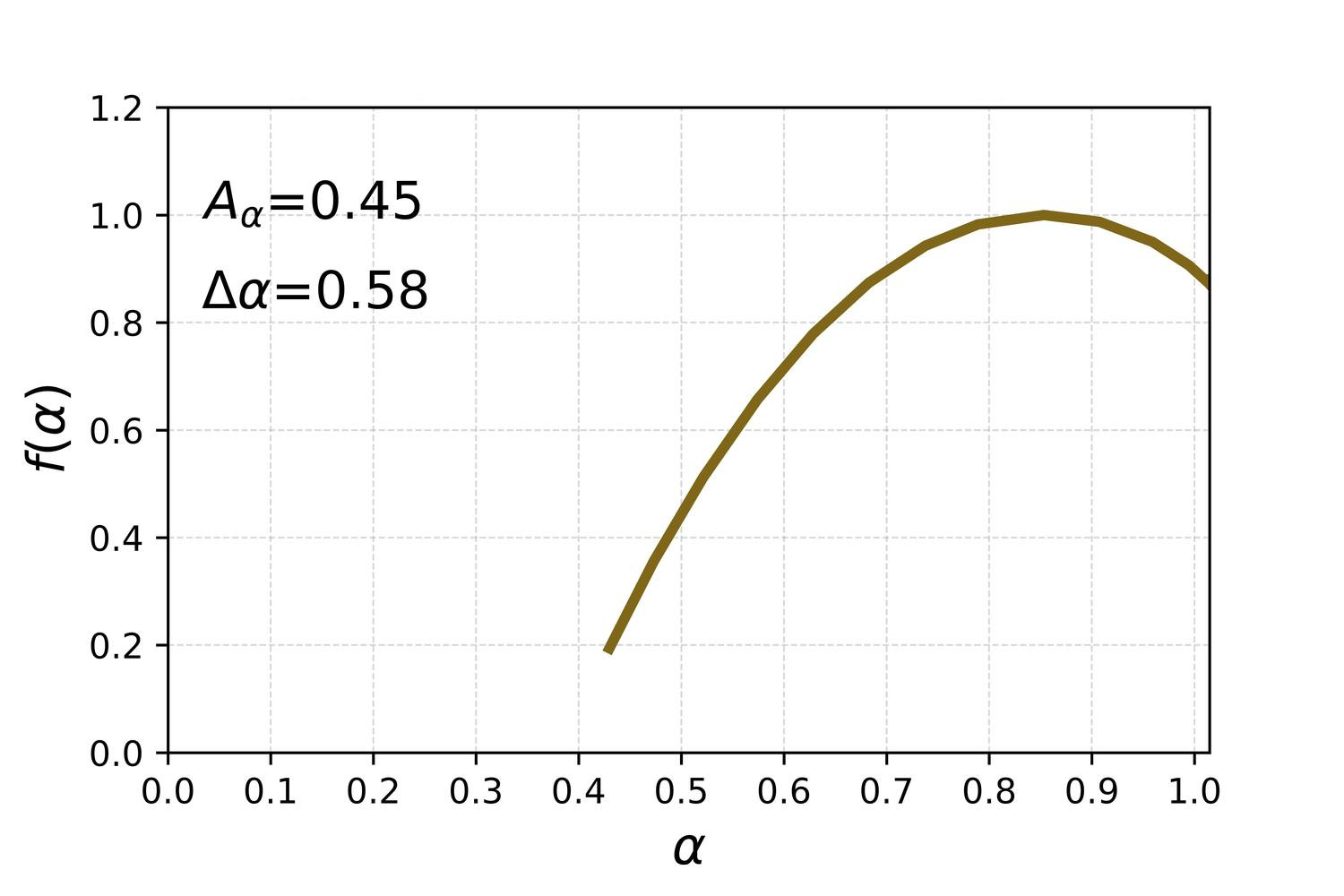}} \\
\cline{2-2}
& \adjustbox{valign=t}{\includegraphics[trim={0cm 0 0 3cm}, clip, width=1.1\linewidth]{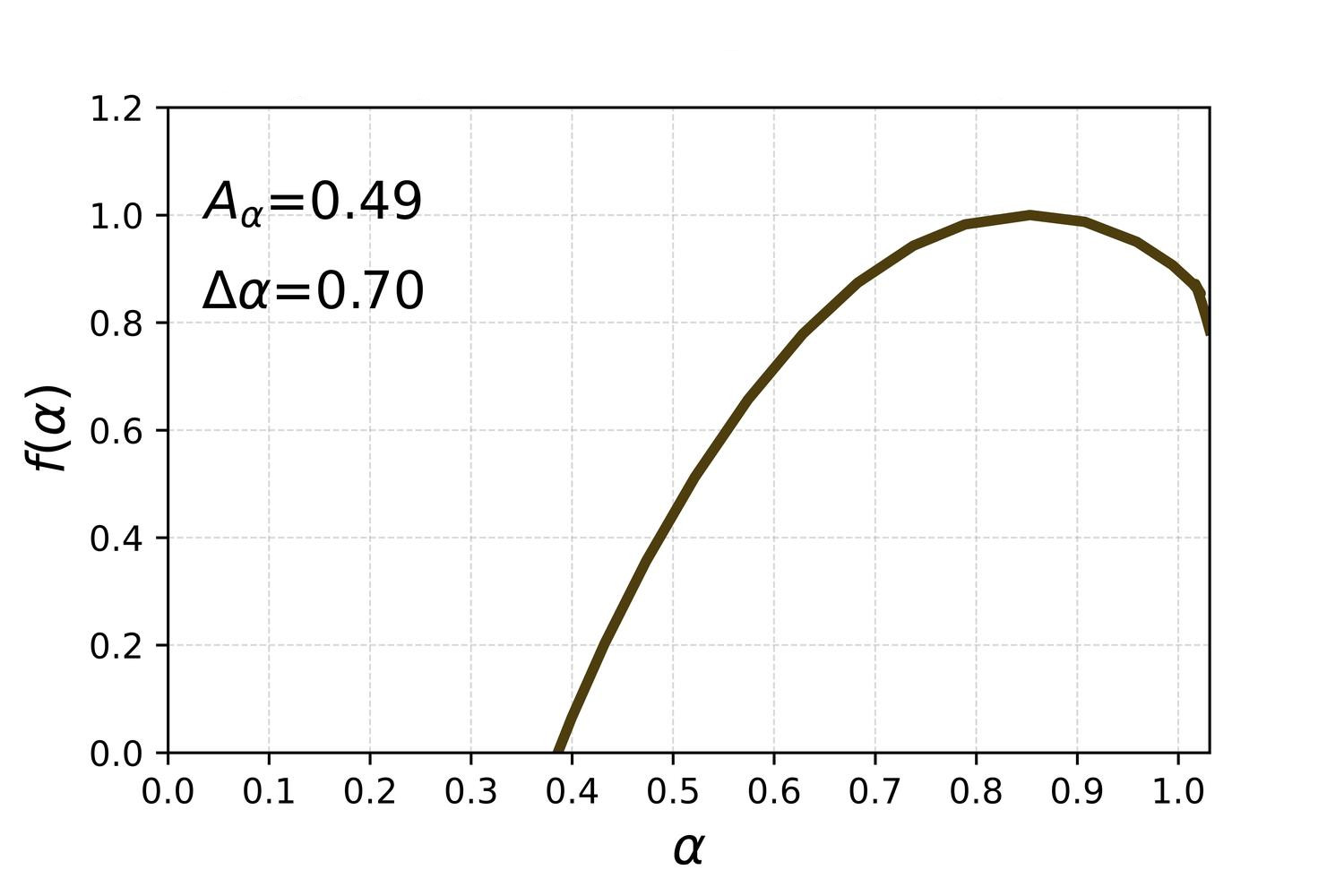}} \\
\hline
\end{tabularx}
\begin{tabularx}{\textwidth}{| >{\centering\arraybackslash}m{0.75\textwidth} | >{\centering\arraybackslash}m{0.25\textwidth} |}
\hline
\multirow{3}{*}{
\adjustbox{valign=t}{\includegraphics[trim={0cm 0 0 3cm}, clip, width=1.07\linewidth]{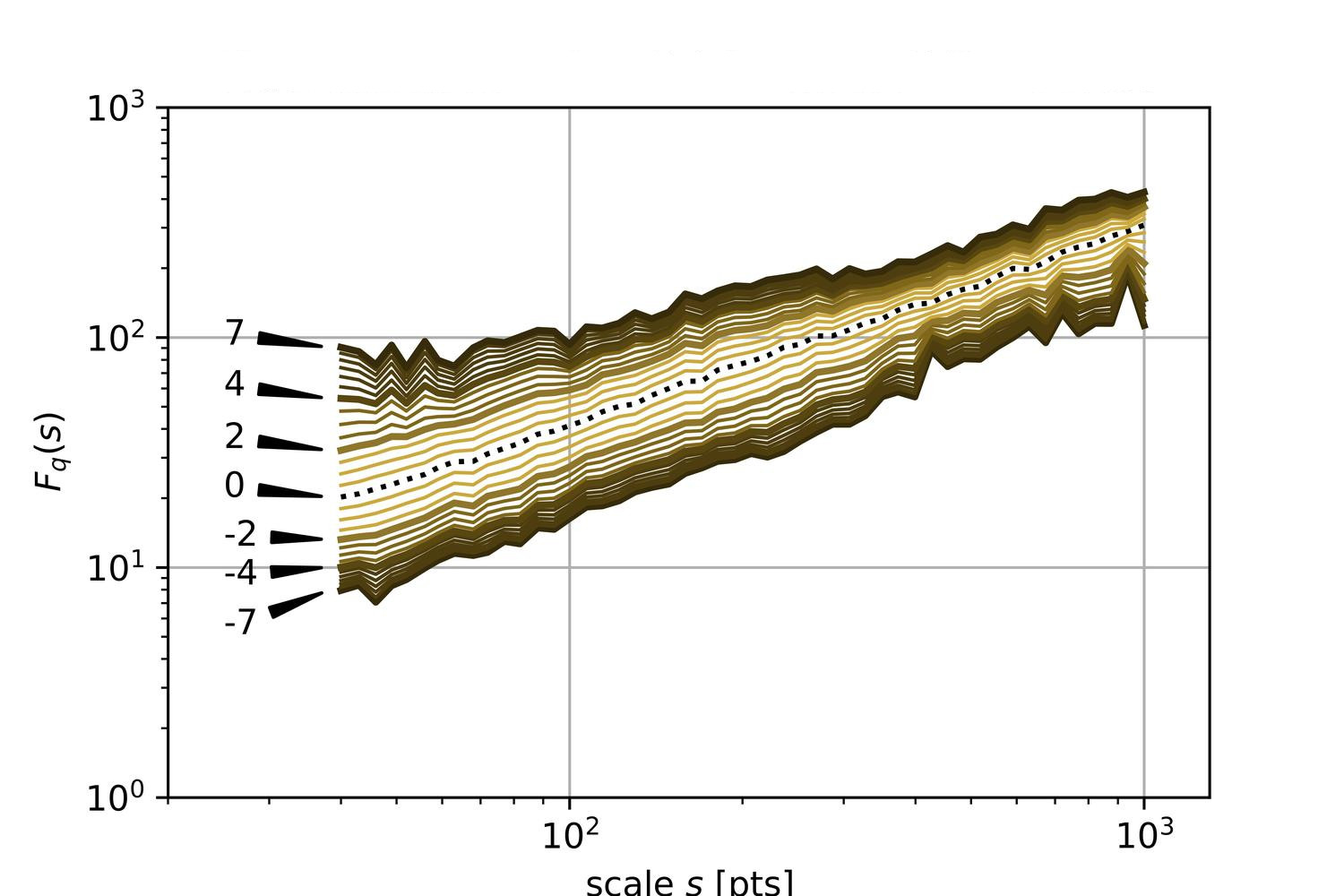}}
}
& \adjustbox{valign=t}{\includegraphics[trim={0cm 0 0 3cm}, clip, width=1.1\linewidth]{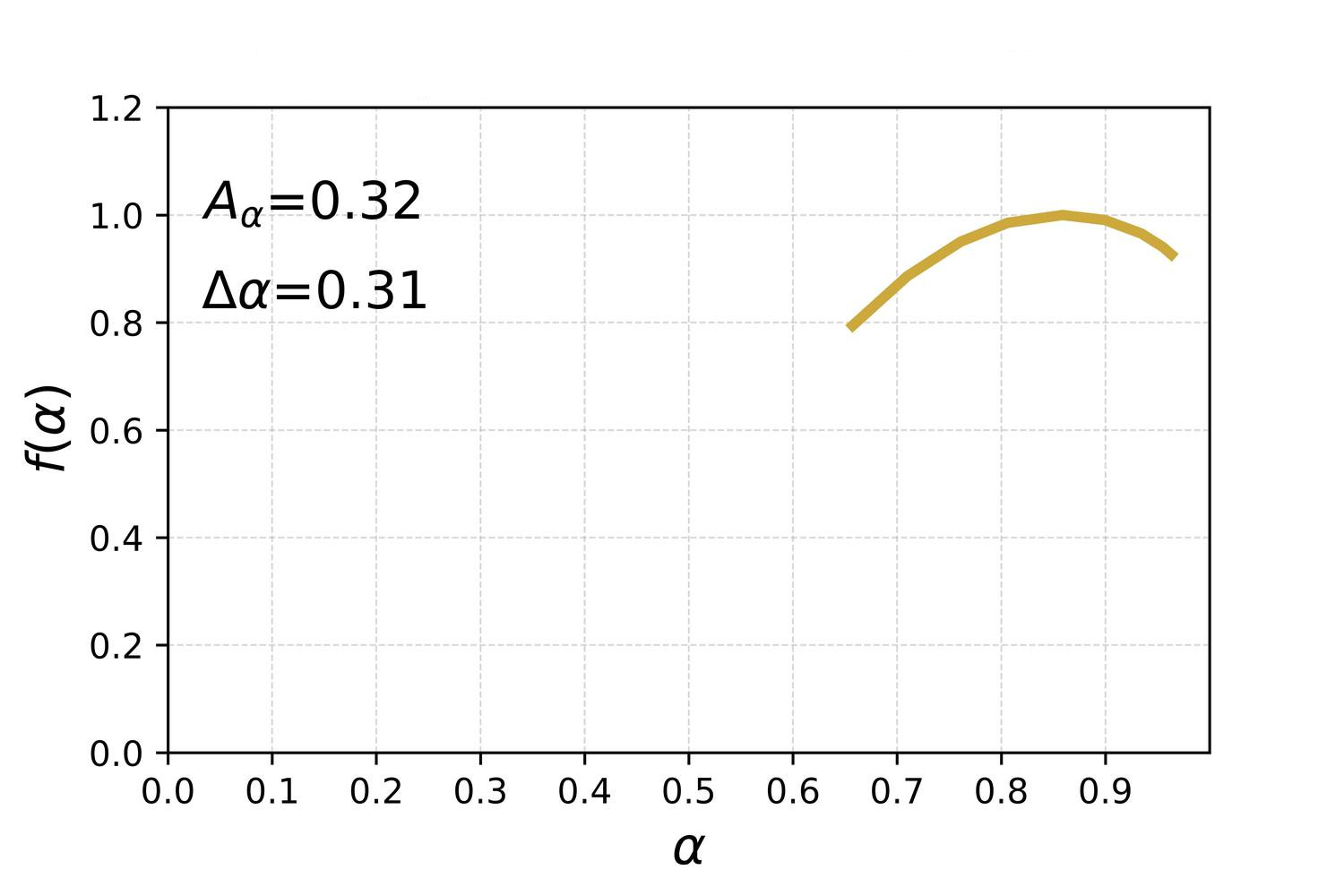}} \\
\cline{2-2}
& \adjustbox{valign=t}{\includegraphics[trim={0cm 0 0 3cm}, clip, width=1.1\linewidth]{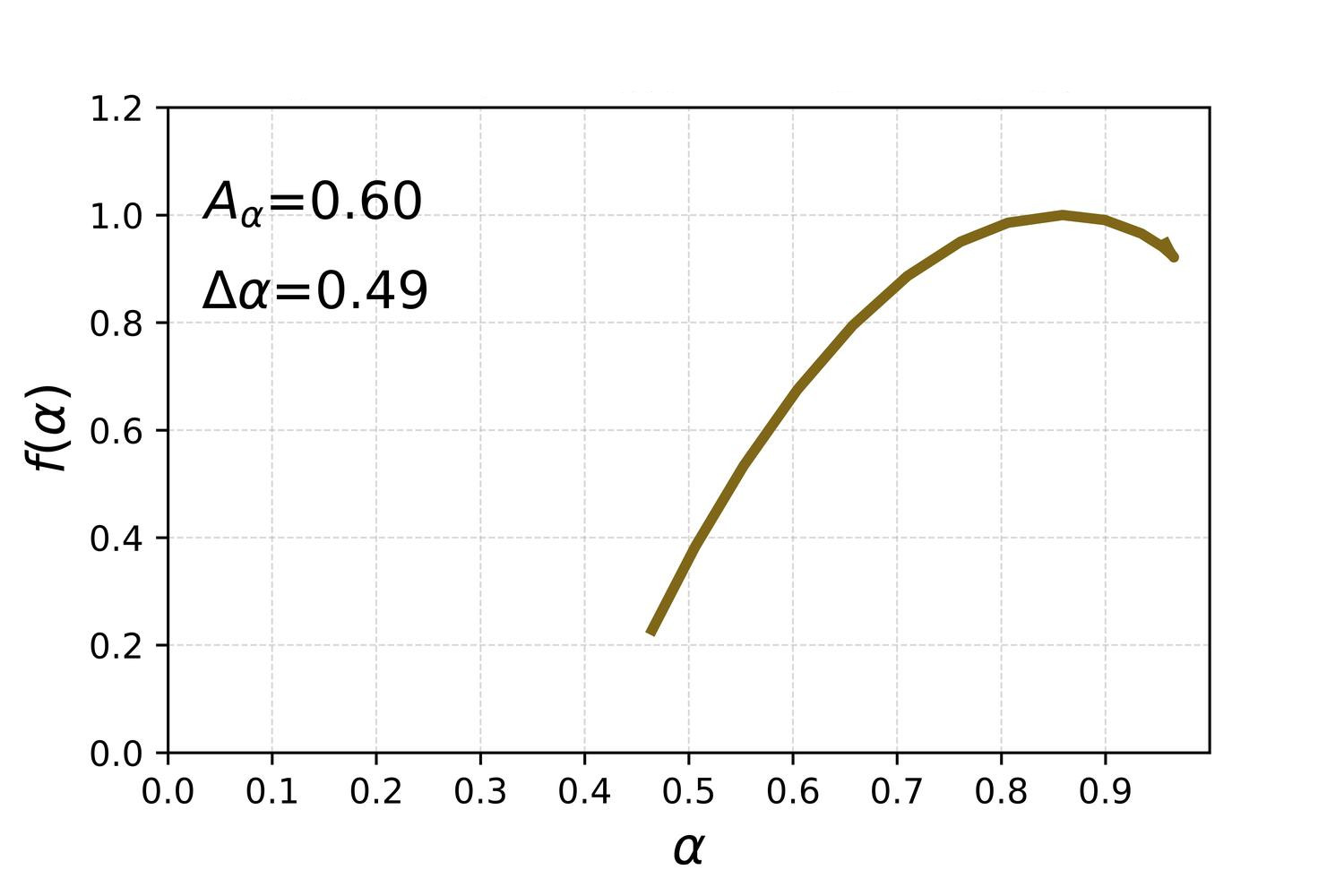}} \\
\cline{2-2}
& \adjustbox{valign=t}{\includegraphics[trim={0cm 0 0 3cm}, clip, width=1.1\linewidth]{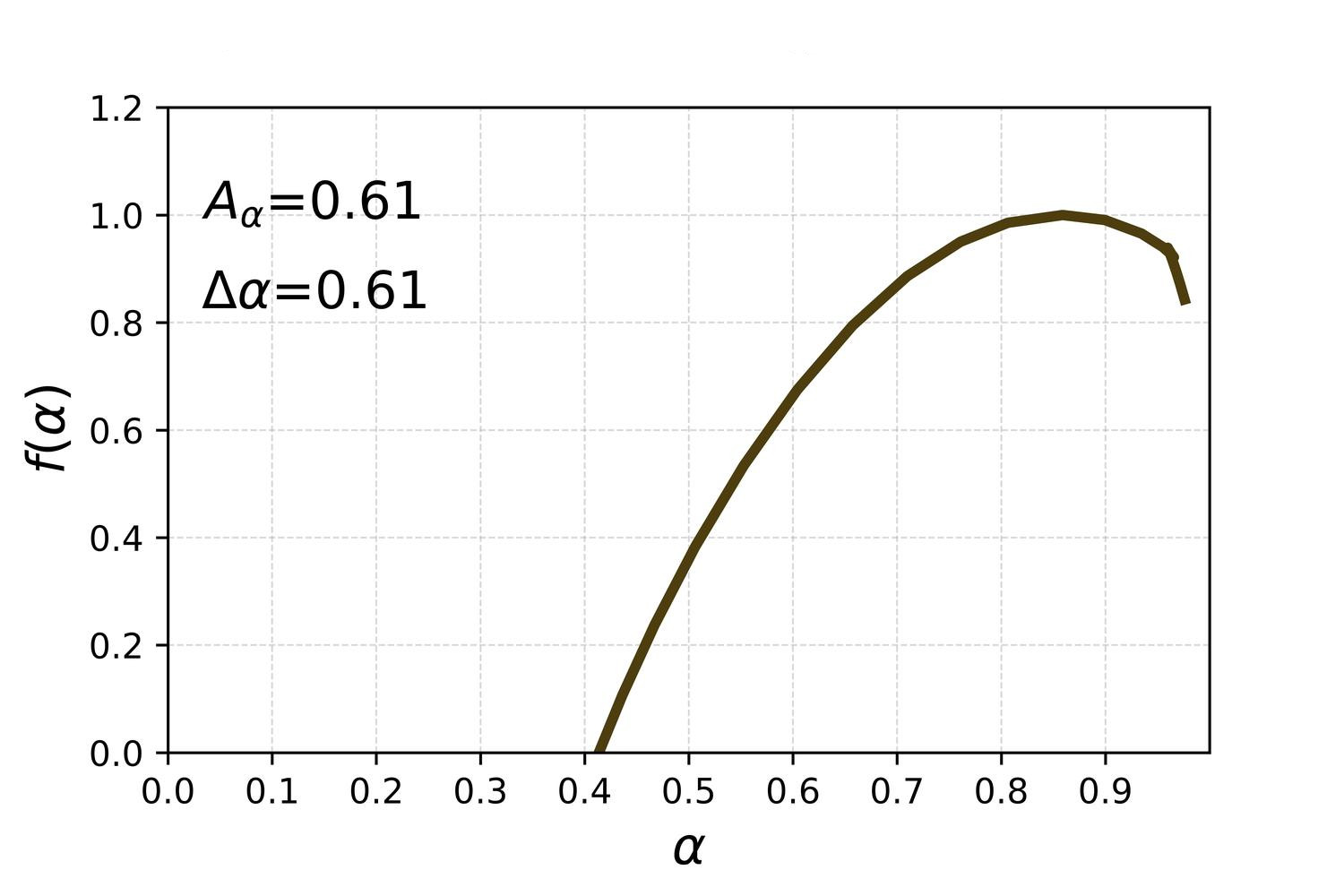}} \\
\hline
\end{tabularx}
\centering
\includegraphics[width=0.9\linewidth]{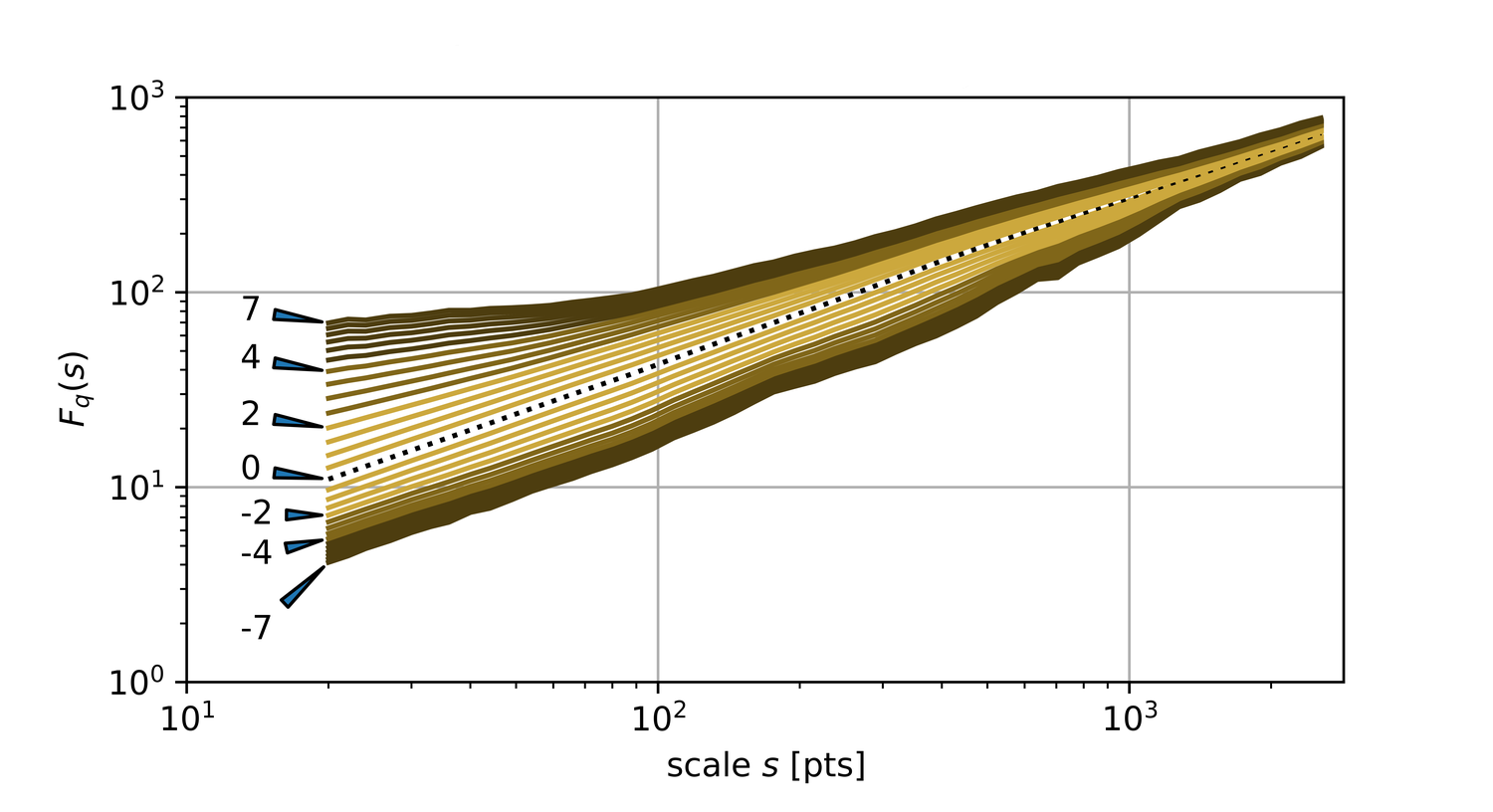}
\end{adjustwidth}
\caption{The same functions as in Figure~\ref{fig::slv.Fq.printed} for the Spanish version of \textit{Hopscotch} with two random permutations of chapters (\textbf{top} and \textbf{middle}) and the average taken over a set of 1000 sample permutations (\textbf{bottom}).}
\label{fig::slv.Fq.random}
\end{figure}

The family of the generalised Hurst exponents $h(q)$ is closely related to the H\"older exponents $\alpha$ and the singularity spectrum $f(\alpha)$ via Equation~(\ref{eq::singularity.spectrum}). From the perspective of the amount of information that one can extract from the data, the $f(\alpha)$ representation is more natural and convenient than $h(q)$, especially if a fractal analysis is carried out. However, both $h(q)$ and $f(\alpha)$ are of little use without inspecting of the fluctuation functions $F_q(s)$ first. This is because they carry information about the possible existence of fractal scaling over a sufficient range of scales $s$, which is a crucial characteristic of the data. Therefore, \mbox{Figures~\ref{fig::slv.Fq.printed}--\ref{fig::slv.Fq.random}} show the family of $F_q(s)$ calculated in the range $-7 \le q \le 7$ for different languages and different chapter orders.

Let one start with the printed order of chapters; the corresponding fluctuation functions for different values of $q$ are displayed in Figure~\ref{fig::slv.Fq.printed} (main panels) for the three languages considered in this work. All the functions show a power-law dependence (approximately straight lines on double logarithmic plots) over some range of scales, whose right limit reaches typically 300--400 consecutive sentences for both positive and negative $q$s. The singularity spectra derived from the scaling exponents $h(q)$ for different intervals of $q$ are shown in the side panels of each main panel of Figure~\ref{fig::slv.Fq.printed}. The spectra have their maximum at $\alpha\approx 0.8$, which points out to a strong persistence in SLV. By going through these side panels from top to bottom, the range of the values of $q$ gradually extends from $-2 \le q \le 2$ to $-7 \le q \le 7$. The rationale behind considering different spans of $q$ is that there is a certain degree of asymmetry in the density of the lines related to the functions $F_q(s)$ between positive and negative $q$s in the main panels. Indeed, if one looks at the values of the asymmetry index $A_{\alpha}$ for the $f(\alpha)$ spectra, it assumes a minimum value for the most narrow range of $q$ and it increases with the increasing $|q|$. This observation is valid for all the three languages even though particular values of $A_{\alpha}$ can differ from each other for the equivalent intervals of $q$ in different languages. The asymmetry is left-sided, which means that the multiscaling comes predominantly from large fluctuations in SLV. This type of asymmetry is more often seen in empirical data than their opposite counterpart~\cite{DrozdzS-2015a}. As regards the widths $\Delta \alpha$ of the $f(\alpha)$ spectra, they increase with $|q|$, which is a normal behaviour in such a context, but their value indicates that the SLV time series under study show the multifractal scaling even for the most narrow range of $q$. This result has been expected, since one of our past studies focused on books with a stream-of-consciousness narrative~\cite{DrozdzS-2016a}.

Now, the order of chapters can be altered to agree with the one recommended by Cort\'azar. This is his ``hopscotch'' order, which allows a reader for viewing the book's plot from a different perspective. The respective results in terms of $F_q(s)$ and $f(\alpha)$ are documented in Figure~\ref{fig::slv.Fq.recommended}, whose structure is the same as the structure of Figure~\ref{fig::slv.Fq.printed}. By altering the order, one destroys the temporal structure of the SLV time series for scales above the average length of a chapter expressed in the number of sentences, while the structure on shorter time scales remains largely unaltered. This leads to a visible significant distortion of the behaviour of $F_q(s)$ for the scales $s \approx 10^3$ with respect to the printed order of chapters, which tends to be inclining towards monofractality there (a narrow beam of almost parallel lines). In contrast, for the scales inside the chapters, the changes are less evident. By looking at the singularity spectra obtained for the recommended order, one observes a systematic increase in both the asymmetry $A_{\alpha}$ and the width $\Delta \alpha$ if compared to the results for the printed order. This effect suggests an existence of a richer variety of the singularity strengths expressed by the H\"older exponents $\alpha$ in the considered time series. The origin of this behaviour remains unclear at the present stage of research, however. Together with the increased left--right asymmetry of the spectra, a slight shift of the maximum of $f(\alpha)$ in the direction of larger values of $\alpha$ can be identified for the recommended order ($\alpha \approx 0.85$). This seems to be a systematic effect for the three languages. Finally, the change in $A_{\alpha}$ is related to this shift, because together with the unchanged value of $\alpha_{\rm min}$ they cause the elongation of the left branch of $f(\alpha)$, while the length of the right branch remains roughly the same in both situations despite the fact that $\alpha_{\rm max}$ is larger for the recommended order than it is for the printed order.

It is interesting to investigate whether such effects can also be observed for any random permutation of chapters. Figure~\ref{fig::slv.Fq.random} shows the results of the multifractal analysis of the SLV data corresponding to such permutations. The results for two individual permutations are shown there together with the average results for 1000 different random permutations. Only the results for the original Spanish version of the text are shown; the remaining two language versions are characterised by quantitatively similar results. Randomisation of the chapter order brings results that do not differ much from the results for the recommended order: for both individual permutations, the spectrum width $\Delta \alpha$ is elevated with respect to its value for the printed order (0.65 and 0.70 vs. 0.56) but it is comparable to the value for the recommended order (0.65 and 0.70 vs. 0.65). Regarding the asymmetry index $A_{\alpha}$, the situation is not so evident as in the case of $\Delta \alpha$, because the results for one of the random permutations and for the printed order are close to each other (0.49 vs. 0.46), even though a larger discrepancy is also possible (0.61 vs. 0.46). These numbers refer to the maximum span of $q$ ($-7 \le q \le 7$). It is also instructive to compare the results for individual chapter orders with the average over many random permutations (see Figure~\ref{fig::slv.Fq.random} (bottom)).

The significance of the results reported above has also been tested against the null hypothesis of no correlation. This has been performed by a conventional surrogate testing~\cite{TheilerJD-1996a}, in which the SLV time series are randomised at the individual sentence level, which destroys memory. Results show that in this case the singularity spectra are shifted to smaller values of $\alpha$ and the maximum of $f(\alpha)$ is located near $\alpha=0.5$. Even though $\Delta \alpha$ for such surrogate data remains as high as 0.2, the inherent properties of the MFDFA procedure suggest that $f(\alpha)$ has been broadened spuriously by the finite-size effects related to the heavy-tailed probability distribution functions of the SLV data~\cite{DrozdzS-2009a,KwapienJ-2023a}. Another surrogate testing based on the Fourier-phase randomisation of time series, which destroys all nonlinear correlations~\cite{SchreiberT-1996a}, brings the expected results, i.e., the singularity spectra become point-like and located exactly at $\alpha \approx 0.5$. Both types of surrogate tests confirm the statistical validity of the presented outcomes.

\section{Summary}

The time series of SLV representing Julio Cort\'azar's novel \textit{Hopscotch} in its original Spanish version and in two foreign-language translations were analysed by means of the autocorrelation function, the Weibull analysis, and the MFDFA. Long-range memory effects were observed in terms of a power-law decay of ACF that extended beyond the scales defined by the average chapter length for the printed order of chapters and for other possible chapter orders, including the one recommended by the author. The fluctuations of the SLV time series were distributed in agreement with an exponential distribution, at least to a certain extent (small fluctuations somehow deviated from this pattern). For a comparison, the time series of PMDV, where all punctuation marks were considered, were modelled by the discrete Weibull distribution with $\beta>1$---the value that did not differ from its standard values for written texts. Finally, by using the MFDFA formalism, the fractal properties of the SLV time series were studied. They happened to be multifractal for both the printed and recommended orders, as well as for all the considered random orders. The main difference between the results obtained for different chapter orderings was the observation that all the conceptual orders (i.e., the non-printed ones) showed an enhanced variety of singularities (i.e., broader singularity spectra) present in the time series and expressed by the H\"older exponents. They also developed stronger left-sided asymmetry as compared with their counterparts for the printed order of chapters, the effect produced by a subtle but noticeable shift of the maxima of the spectra towards larger values of the H\"older exponent. It has to be stressed that multifractal structures in literary works are a rare property that have been identified only in a small fraction of texts~\cite{DrozdzS-2016a,StaniszT-2024a}. Interestingly, this and the other considered properties seem to be invariant under translation into foreign languages if a translator pays sufficient attention to the structure of the original (see also~\cite{AusloosM-2010a,StaniszT-2023a,DecJ-2024a}).

The results obtained in this analysis indicate that the printed order of chapters in \textit{Hopscotch} cannot be distinguished statistically if compared to either the recommended order or a generic order obtained via a random permutation of chapters. Also, we do not observe any qualitative difference between the recommended order and generic orders, which leads us to a conclusion that neither the printed order nor the recommended order introduce significant temporal correlations on scales above the average number of sentences in a chapter, at least such correlations that could be identified with the methodology applied here. From this perspective, each chapter constitutes a largely independent block whose position in the narrative can be arbitrarily changed without changing the statistical properties of SLV.

If the results of this study are compared with the earlier results for \textit{Finnegans Wake} by J. Joyce, one observes that \textit{Hopscotch} is more conventional not only in terms of literary style but also in terms of the statistical properties~\cite{DrozdzS-2016a,StaniszT-2024a,StaniszT-2024b}. Its role in the world's literature stems from its innovative construction including the nonlinear and reader-engaging narrative rather than the structural and lingual complexity that is so striking in \textit{Finnegans Wake} or \textit{Ulysses}. Nevertheless, the results of this work provide an inspiration for more systematic research of the most prominent and unique literary works, as these are the ones that are potent to catalyze opening new perspectives in linguistics, including those in the field of large language models. {In the latter context, it is important that the methodology presented here allows for the identification and quantification of various types of correlations in written texts, including long-range ones. In the terminology of nonlinear dynamics this means a reduction in the effective dimensionality of the corresponding `phase space' and therefore a reduction in the actual number of parameters involved. In LLMs, nowadays commonly based on neural networks, such facts can be exploited to significantly, and perhaps even crucially, reduce the number of applied parameters and thus increase their efficiency in terms of time and energy consumption. 

A related prospective direction of future research, somehow parallel yet substantially distinct from the multifractal approach discussed here, is a dynamically oriented analysis of SLV, in which one looks for low-dimensional attractors in a reconstructed phase-space of ``text-writing dynamics''. Apart from some early attempts~\cite{KosmidisK-2006a,AusloosM-2012a}, this direction still remains largely unexplored.}
\vspace{6pt} 

\authorcontributions{Conceptualization, J.D., M.D., S.D. and T.S.; Methodology, S.D. and J.K.; Software, J.D., M.D., J.K. and T.S.; Validation, J.D., M.D., S.D., J.K. and T.S.; Formal analysis, J.D., M.D., S.D. and J.K.; Investigation, J.D., M.D., S.D., J.K. and T.S.; Resources, J.D., M.D. and T.S.; Data curation, J.D., M.D. and T.S.; Writing---original draft, J.K.; Writing---review \& editing, S.D., J.K. and T.S.; Visualization, J.D. and M.D.; Supervision, S.D.}

\funding{This research received no external funding.}

\institutionalreview{Not applicable.}

\dataavailability{The data presented in this study are available on request from the corresponding author. The data are not publicly available due to copyright.}

\conflictsofinterest{The authors declare no conflicts of interest.}

\begin{adjustwidth}{-\extralength}{0cm}
\reftitle{References}

\PublishersNote{}
\end{adjustwidth}
\end{document}